\documentclass[10pt,journal,compsoc]{IEEEtran}
\ifCLASSOPTIONcompsoc
  \usepackage[nocompress]{cite}
\else
  \usepackage{cite}
\fi
\ifCLASSINFOpdf
   \usepackage[pdftex]{graphicx}
\else
\fi
\usepackage{amsmath}
\usepackage{multirow}
\usepackage{booktabs}
\usepackage{threeparttable}
\usepackage{hyperref}
\hypersetup{
    colorlinks=true,
    linkcolor=black,
    filecolor=blue,      
    urlcolor=blue,
    citecolor=black,
}

\usepackage{array}
\begin{document}
%
\title{Revisiting Light Field Rendering with Deep Anti-Aliasing Neural Network}
%
%
%
%
\author{Gaochang~Wu,~
        Yebin~Liu,~\IEEEmembership{Member,~IEEE,}
				Lu~Fang,~\IEEEmembership{Senior Member,~IEEE,} 
        and~Tianyou~Chai,~\IEEEmembership{Fellow,~IEEE}
\IEEEcompsocitemizethanks{\IEEEcompsocthanksitem Gaochang Wu and Tianyou Chai are with the State Key Laboratory of Synthetical Automation for Process Industries, Northeastern University, Shenyang 110819, China, and also with the Institute of Industrial Artificial Intelligence, Northeastern University, Shenyang 110819, China. E-mail: \{wugc, tychai\}@mail.neu.edu.cn.
\IEEEcompsocthanksitem Yebin Liu is with Department of Automation, Tsinghua University, Beijing 100084, P. R. China. E-mail: liuyebin@mail.tsinghua.edu.cn.
\IEEEcompsocthanksitem Lu Fang is with the Department of Electronic Engineering, Tsinghua University, Beijing 100084, China, and also with the Beijing National Research Center for Information Science and Technology, Beijing 100084, China. E-mail: fanglu@tsinghua.edu.cn.}

\thanks{Corresponding Author: Tianyou Chai.}
}

%
%

\markboth{}%
{Shell \MakeLowercase{\textit{et al.}}: Bare Demo of IEEEtran.cls for Computer Society Journals}
%



\IEEEtitleabstractindextext{%
\begin{abstract}
The light field (LF) reconstruction is mainly confronted with two challenges, large disparity and the non-Lambertian effect. Typical approaches either address the large disparity challenge using depth estimation followed by view synthesis or eschew explicit depth information to enable non-Lambertian rendering, but rarely solve both challenges in a unified framework. In this paper, we revisit the classic LF rendering framework to address both challenges by incorporating it with advanced deep learning techniques. First, we analytically show that the essential issue behind the large disparity and non-Lambertian challenges is the aliasing problem. Classic LF rendering approaches typically mitigate the aliasing with a reconstruction filter in the Fourier domain, which is, however, intractable to implement within a deep learning pipeline. Instead, we introduce an alternative framework to perform anti-aliasing reconstruction in the image domain and analytically show comparable efficacy on the aliasing issue. To explore the full potential, we then embed the anti-aliasing framework into a deep neural network through the design of an integrated architecture and trainable parameters. The network is trained through end-to-end optimization using a peculiar training set, including regular LFs and unstructured LFs. The proposed deep learning pipeline shows a substantial superiority in solving both the large disparity and the non-Lambertian challenges compared with other state-of-the-art approaches. In addition to the view interpolation for an LF, we also show that the proposed pipeline also benefits light field view extrapolation.
\end{abstract}

\begin{IEEEkeywords}
Light field reconstruction, light field rendering, deep learning, view extrapolation.
\end{IEEEkeywords}}

\maketitle

\IEEEdisplaynontitleabstractindextext

%
\IEEEpeerreviewmaketitle

\IEEEraisesectionheading{\section{Introduction}}\label{sec:introduction}

%
%
%
%
\IEEEPARstart{A}{s} an alternative to traditional 3D scene representation using scene geometry (or depth) and texture (or reflectance), light field (LF) achieves photorealistic view synthesis in real-time using LF rendering technology~\cite{LFrendering,gortler1996lumigraph}. This high-quality rendering requires the disparities between adjacent views to be less than one pixel, i.e., the so-called densely-sampled LF. Unfortunately, the practical scenarios including dynamic scene~\cite{CameraArray} or limited acquisition time~\cite{LFrig} impose insufficient sampling in the angular dimension. The quality of the rendered novel views is inevitably perturbed by the large disparity (range) in the sampled LF. In addition, the potential non-Lambertian effect in the scene, such as jewellery, fur, glass and face, will further aggravate this side effect~\cite{chai2000plenoptic,zhang2003spectral}.

\begin{figure}
\begin{center}
\includegraphics[width=1\linewidth]{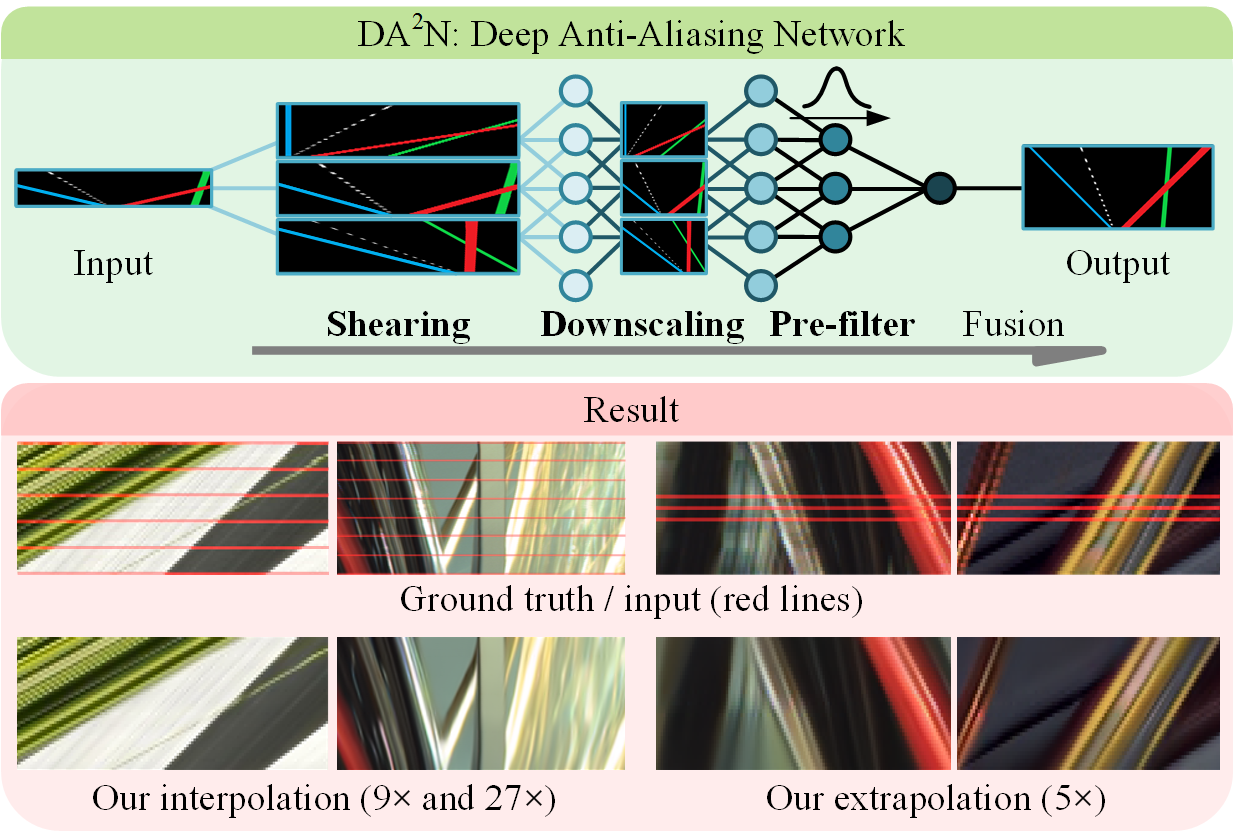}
\end{center}
\vspace{-6mm}
   \caption{Schematic diagram of the proposed Deep Anti-Aliasing Network (DA$^2$N) with embedding shearing, downscaling and prefiltering. In the result, we use input views indicated by the red lines for the EPI reconstruction, including interpolation and extrapolation.}
\label{fig:Teaser}
\vspace{-4mm}
\end{figure}

\begin{figure*}
\begin{center}
\includegraphics[width=1\linewidth]{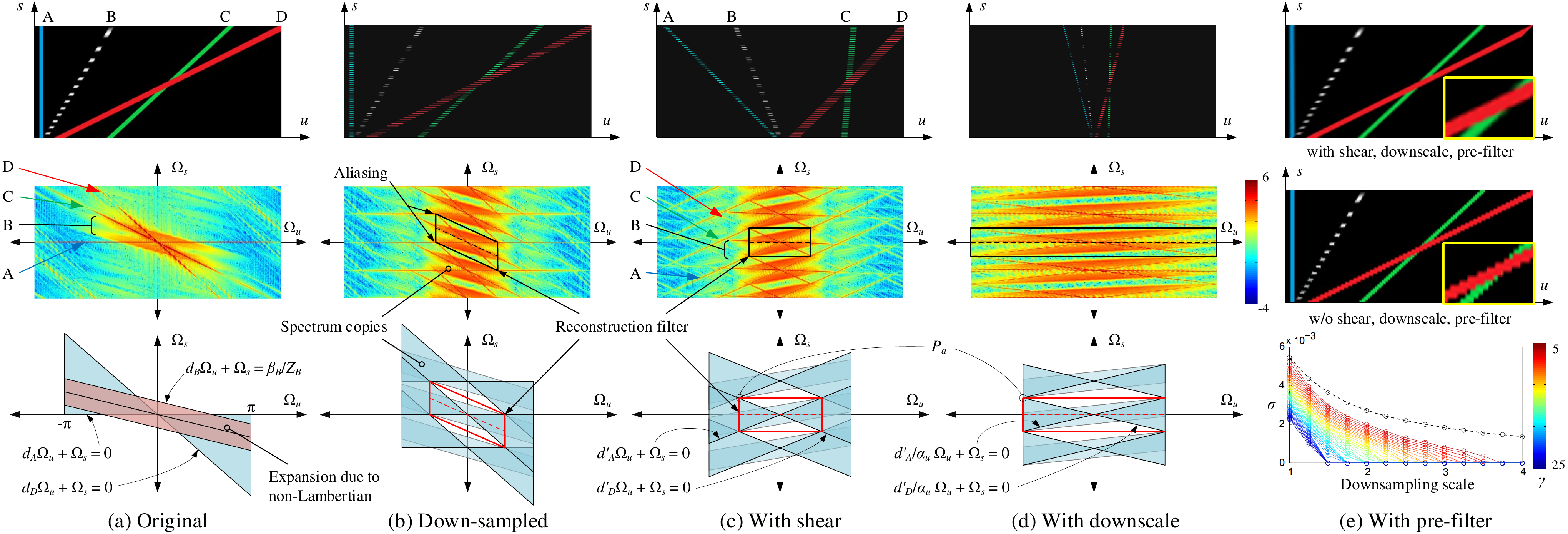}
\end{center}
\vspace{-6mm}
\caption{Fourier analysis of shear, downscaling (spatial) and prefilter operations. The EPI for demonstration is composed of three Lambertian points ($A, C$ and $D$) and one non-Lambertian point ($B$) with different disparities. (a) Original EPI (top), its Fourier transform after taking the logarithm (middle) and diagram (down). The non-Lambertian point will introduce expansionary spectra on each side of the $\Omega_s$ axis \cite{zhang2003spectral}; (b) Spectrum replicas appear when the EPI is angularly down-sampled. The overlapped spectrum and its replicas will lead to an aliasing effect. An ideal reconstruction filter (parallelogram) can be obtained using the optimal rendering disparity (depth) shown by the dashed line; (c) Shearing the EPI with a proper value forces the optimal rendering disparity of the reconstruction filter to rotate towards the $\Omega_u$ axis; (d) With a downscaling (spatial) operation, the original spectrum will shrink towards the $\Omega_u$ axis. And the separated spectra substantially mitigates the aliasing issue; (e) We show an EPI with shear, downscaling and prefilter operations (top) and an EPI with directly up-sampling (middle), see the close-up for the detail. We can efficiently decrease the shape parameter $\sigma$ of the prefilter by using shear and downscaling operations (bottom).}
\label{fig:FA}
\vspace{-3mm}
\end{figure*}

To overcome these two technical challenges, i.e., large disparity and non-Lambertian effect, classic LF rendering approaches~\cite{chai2000plenoptic,zhang2003spectral,Shearlet} tended to treat the plenoptic sampling and reconstruction in a signal processing way. They may design a powerful anti-aliasing filter to reconstruct the LF in the Fourier domain, as illustrated in Fig. \ref{fig:FA}(b). However, it relies on a carefully designed filter determined by the depth information and the aliasing degree, which are prone to blurry results due to the high-frequency loss~\cite{stewart2003new}. With the boost of tremendous deep learning techniques, state-of-the-art approaches achieve a promising result, either depth estimation followed by view synthesis~\cite{DeepStereo,DoubleCNN} or direct LF reconstruction~\cite{LFCNN,WuEPICNN2018,YeungECCV2018}. However, existing view synthesis or LF reconstruction methods can hardly solve the aforementioned challenges together. For instance, Kalantari~\textit{et al}.~\cite{DoubleCNN} developed sequential networks to infer the scene depth and synthesize novel view via the four corner sub-aperture images. Since the depth estimation is usually based on the Lambertian assumption, it will lead to an ambiguous result in a non-Lambertian surface (see Fig. \ref{fig:Result3}). Alternatively, some other approaches~\cite{WuEPICNN2018,YeungECCV2018} eschew explicit depth information to enable non-Lambertian rendering, while they suffer from either aliasing or blurring effect under the large disparity (see Fig. \ref{fig:Result2} and Fig. \ref{fig:Result3}). It is apparent that the effect caused by either the large disparity or the non-Lambertian effect is ubiquitous in the real-world, neglecting either of them will certainly lead to quality degeneration in the reconstructed LF.

In this paper, we consider the two challenges, large disparity and non-Lambertian effect, in a unified deep learning framework. The detailed analysis in the Fourier domain reveals that the essential issue behind the two challenge is the aliasing problem~\cite{chai2000plenoptic,zhang2003spectral}. The intuitive way of mimicking the anti-aliasing filter~\cite{stewart2003new,Shearlet} via a neural network in the Fourier domain is intractable due to the requirement of a global spectrogram perception (Sec. \ref{Sec:Problem1}). Our insight is decomposing the classical anti-aliasing filtering in the \textit{Fourier domain} into a sequence of operations, including ``shearing, downscaling and prefiltering'' in the \textit{image domain}, and formulate them with the deep learning pipeline. Compared with other available learning-based methods that resort to a deeper representation, our shearing-downscaling-prefiltering framework owns the advantage of explicitly and efficiently aliasing handling from the essential structure design, and thus, is more competitive in solving large disparity and non-Lambertian scenarios.

Specifically, the shear operation~\cite{Tao, ren05} converts the anti-aliasing filter with a complicated representation in the Fourier domain into a simple low-pass filter (i.e., the Gaussian filter) in the image domain via plane (sub-aperture image) sweep. Subsequently, rather than directly applying the Gaussian filter, we address the aliasing issue by combining downscaling and prefiltering in the image domain. The downscaling operation indirectly reduces the disparity range~\cite{chai2000plenoptic} by resizing the spatial resolutions of the input LF. And the prefiltering operation further suppresses the aliasing high-frequencies introduced by the non-Lambertian effect. The Fourier analysis indicates that the proposed shearing-downscaling-prefiltering framework is comparable to the conventional reconstruction filter in the Fourier domain.

With the guidance of the sequential operations, the designed network is able to explicitly and efficiently solve both challenges while also reasoning in an end-to-end manner. We achieve this through the following features. First, we apply a multi-branch structure to implement shear operation with different candidates. Accordingly, a fusion network  is employed to post-fuse the reconstructed LFs with different shear amounts and produce the high angular resolution LF without depth estimation. Second, we propose a trainable Laplacian decomposition to formulate the downscaling operation and keep the high-frequencies simultaneously. Third, we use a convolutional layer in the spatial dimension to perform prefilter operation, which learns to handle the aliasing high-frequencies from training data. The overall network is shown in Fig. \ref{fig:Teaser}, and is termed as Deep Anti-Aliasing Network, or DA$^2$N for short.

To further boost the performance, we begin by training the DA$^2$N from scratch on regular LFs and then fine-tune the network using pseudo epipolar plane images (EPIs) from unstructured LFs (Sec. \ref{Sec:dataset}). On the one hand, we indicate that pseudo EPIs and regular EPIs with the non-Lambertian effect not just have the same visual features, but more importantly, share the same properties regarding the aliasing issue. On the other hand, unstructured LF is an easily accessible source (e.g., by a hand-held camera) and is able to offer a large disparity compared with a regular LF (e.g., Lytro LF and HCI synthetic LF \cite{HCI}). The technical contributions are summarized as follows:
\begin{itemize}
    \item We propose a novel end-to-end Deep Anti-aliasing Neural Network (DA$^2$N) to solve the two challenges, large disparity and non-Lambertian effect, in light field rendering. For the network training, the pseudo EPIs from unstructured LFs are particularly used.
    \item We provide a theoretical analysis of the aliasing problem that leads to the two challenges. And the proposed ``shearing, downscaling and prefiltering'' in the image domain is comparable to the conventional reconstruction filter in the Fourier domain.
    \item We demonstrate the superiority of the DA$^2$N for high-quality LF reconstruction with both non-Lambertian effect and large disparity (up to 14 pixels), as shown in Fig. \ref{fig:Result1_2}. In addition, by combining with a discriminator, DA$^2$N can be further extended to view extrapolation, as shown in Fig. \ref{fig:Teaser}.
\end{itemize}

\section{Related Work}
\textit{Depth-based view synthesis.} This kind of approaches typically first estimate the depth of a scene~\cite{Tao,Occ,sulc2017inverse,schilling2018trust,alperovich2018light}, such as structure tensor-based local direction estimation in the epipolar plane image (EPI) domain~\cite{Wanner}, and phase-based depth estimation in the sub-aperture image domain~\cite{binolf}. Then the input views are warped to novel viewpoints and blended in different manners, e.g., soft blending~\cite{soft3D} and learning-based synthesis~\cite{Zheng2018ECCV}. In recent years, some deep learning based methods for directly maximizing the quality of synthetic views have been presented. Flynn \textit{et al.}~\cite{DeepStereo} proposed a deep learning method to synthesize novel views using a sequence of images with wide baselines. Kalantari \textit{et al.}~\cite{DoubleCNN} used two sequential convolutional neural networks to model depth and color estimation simultaneously by minimizing the error between synthetic views and ground truth images. Zhou \textit{et al.}~\cite{zhou2018MPI} trained a network that infers alpha and multiplane images. The novel views are synthesized using homography and alpha composition. However, all these approaches are based on the Lambertian assumption without explicitly addressing the non-Lambertian challenge.

\textit{Reconstruction without explicit depth.} The  LF reconstruction without using explicit depth information or scene geometry is based on sampling and consecutive reconstruction of the plenoptic function~\cite{isaksen2000dynamically,shum2000review, LFfourier, wu2017light}. Typical approaches focus on an anti-aliasing reconstruction in the Fourier domain using signal processing techniques, e.g., quadra-linear interpolation~\cite{LFrendering} and depth corrected interpolation~\cite{gortler1996lumigraph}. Initial work by Chai \textit{et al.}~\cite{chai2000plenoptic} derived a plenoptic sampling theory, which indicates that the recovering of the LF signal can be realized by applying an anti-aliasing reconstruction filter bounded within a double-wedge shape (please refer to Fig. \ref{fig:FA}(b)). Zhang and Chen~\cite{zhang2003spectral} further showed that the non-Lambertian effect in the acquired LF will introduce spectra shifting along the angular dimension, and thus, an expansionary reconstruction filter is required for the LF reconstruction. Instead of using a fixed filter pattern, Vagharshakyan \textit{et al.}~\cite{Shearlet} proposed to adopt a composited reconstruction filter packet to remove the high-frequency spectrums that introduce aliasing effects in Fourier domain. For explicit aliasing handling, Levoy and Hanrahan~\cite{LFrendering} propose to use an optical prefiltering method that acts as a 4D low-pass filter. Alternatively, Chai~\textit{et al.}~\cite{chai2000plenoptic} and Stewart~\textit{et al.}~\cite{stewart2003new} further indicated that a computational prefiltering method can be achieved by using a discrete low-pass filter after the LF sampling.

Based on the recent deep learning techniques, some depth-independent reconstruction approaches were proposed~\cite{LFCNN,WuEPICNN2018,YeungECCV2018,farrugia2019light}. Yeung \textit{et al.}~\cite{YeungECCV2018} applied a coarse-to-fine model using a 4D convolutional network for LF reconstruction followed by an encoder network for view refinement. In spite of the large receptive field in the network, the reconstructed LF can also appear aliasing effects in large disparity cases (as shown in Fig. \ref{fig:Result2}). To explicitly addressing the aliasing effects, Wu~\textit{et al}.~\cite{WuEPICNN2018} proposed to apply a prefilter operation before feeding the EPI into the network. However, limited by the postprocessing using non-blind deblur, the approach is prone to reconstruct an LF with a blurring effect (see Fig. \ref{fig:Result1}). By contrast, we model the prefilter operation within the neural network by adopting 1D convolutional layers with different kernel size and design a trainable Laplacian decomposition for the balance between aliasing handling and high-frequency maintaining (please refer to Sec. \ref{Sec:reconstruction} for detail).

\section{Fourier Analysis: The Aliasing Problem}
\label{Sec:Problem}
In this section, we present a Fourier analysis on the large disparity and non-Lambertian challenges based on the plenoptic theory \cite{chai2000plenoptic,zhang2003spectral}. It reveals that the aliasing problem is the essential issue behind those challenges (Sec. \ref{Sec:Problem1}), where the aliasing degree is determined by the highest frequency of the non-Lambertian effect and the largest disparity. We further introduce an effective framework containing shearing, downscaling and prefiltering operations for anti-aliasing LF reconstruction (Sec. \ref{Sec:Problem2}). Here, each operation is accomplished in the image domain, and thus, can be efficiently formulated with a learning-based pipeline.

\subsection{The large disparity and non-Lambertian challenges}\label{Sec:Problem1}
For a 4D LF with two spatial dimensions $(u,v)$ and two angular dimensions $(s,t)$ represented as $L(u,v,s,t)$, an EPI can be acquired by gathering horizontal lines with fixed $v^*$ along a constant camera coordinate $t^*$, denoted as $E_{v^*,t^*}(u,s)$ (or $E_{u^*,s^*}(v,t)$ similarly). For simplicity, we denote the extracted EPI as $E_{HR}$, where $HR$ stands for high angular resolution. The low angular resolution EPI $E_{LR}$ can be considered as its undersampled version, i.e., $E_{LR}=E_{HR} \downarrow$, where $\downarrow$ denotes the downsampling operation in the angular dimension ($s$ or $t$).

Consider a simple scene composed of three Lambertian points ($A, C$ and $D$) and a non-Lambertian point ($B$) located at different depths. Fig. \ref{fig:FA}(a) (top) shows an EPI extracted from an LF of the scene, where the disparities of point $A, B$ and $C$ are within 1 pixel range while $D$ is larger than 1 pixel. The Fourier transform and its simplified diagram are shown in the middle and bottom of Fig. \ref{fig:FA}(a), respectively. The spectral support of the corresponding line with disparity $d$ is defined by $d\Omega_u+\Omega_s=0$, where $\Omega_u$ and $\Omega_s$ denote the frequencies along the $u$ and $s$ dimensions. Zhang and Chen~\cite{zhang2003spectral} concluded that a non-Lambertian point will introduce expansionary spectra on each side of the $\Omega_s$ axis,
\begin{equation}\label{eq:non-lam}
d\Omega_u+\Omega_s=\pm\beta/Z,
\end{equation}
where $\beta$ is the band width of the band-limited signal caused by the non-Lambertian effect \cite{zhang2003spectral}, and $Z$ is the depth of that non-Lambertian point. The region with red in Fig. \ref{fig:FA}(a) (bottom) demonstrates the non-Lambertian effect in the Fourier domain.

Initial work on LF sampling by Chai \textit{et al.} \cite{chai2000plenoptic} inferred that the down-sampling in the angular dimension will lead to spectrum replicas. Namely, the overlapping of the original spectrum and its replicas will lead to aliasing effect, as shown in Fig. \ref{fig:FA}(b). As indicated in \cite{zhang2003spectral} and also by Eqn. \ref{eq:non-lam}, disparity $d$ and the highest frequency of non-Lambertian effect $\beta$ are two independent factors regarding to the issue. In other words, the non-Lambertian effect will also introduce aliasing even when the corresponding disparity is less than 1-pixel, and the larger disparity or the higher band width of the band-limited signal (caused by the non-Lambertian effect) may worsen the aliasing problem. Therefore, the key to solve the large disparity and non-Lambertian challenges in a unified framework is the aliasing handling.

To address the aliasing issue, a reconstruction filter can be designed for reconstructing the high angular resolution EPI, as shown by the parallelogram in Fig. \ref{fig:FA}(b). The reconstruction filter is defined by the optimal rendering disparity $d_{\textrm{opt}}=\frac{d_{\min}+d_{\max}}{2}$, and the cut-off frequency is determined by the aliasing degree, where $d_{\min}$ and $d_{\max}$ are the minimum and maximum disparities. By incorporating the reconstruction filter with deep learning techniques in the Fourier domain, one may design a network to determine the optimal disparity and the cut-off frequency. However, such a procedure requires a global perception of the entire spectrogram, i.e., the receptive field~\cite{luo2016understanding} of the network should cover the entire input image, which inevitably introduces a mass of parameters.

\begin{table}
\caption{Main symbol used in the paper.}
\vspace{-8mm}
\begin{center}
\begin{tabular}{p{0.9cm}p{2.5cm}<{\centering}|p{0.9cm}p{2.6cm}<{\centering}}
\hline
Symbol & Definition & Symbol & Definition \\
\hline
$L$ & 4D Light field & $E$ & 2D EPI \\
$u, v$ & Spatial dimension & $s,t$ & Angular dimension\\
$d$ & Disparity & $Z$ & Depth \\
$\alpha_h$ & Shear amount & $\alpha_u$ & Downscaling factor\\
$\mathcal{F}$ & Fourier transform & $\gamma$ & Spectrum amplitude\\
\hline
Symbol & \multicolumn{3}{c}{Definition} \\
\hline
$\beta$ & \multicolumn{3}{c}{Band width of the signal caused by non-Lambertian effect} \\
$\alpha_s$ & \multicolumn{3}{c}{Upsampling factor in the angular dimension} \\
$f$ & \multicolumn{3}{c}{Functions defined in the image domain} \\
$\phi$ & \multicolumn{3}{c}{Layer in a neural network} \\
$\theta$ & \multicolumn{3}{c}{Trainable parameters in a neural network}\\
\bottomrule
\end{tabular}
\end{center}
\label{table:symbol}
\vspace{-4mm}
\end{table}

\subsection{Framework for anti-aliasing reconstruction}\label{Sec:Problem2}
Instead of adopting a reconstruction filter in the \textit{Fourier domain} as used in \cite{chai2000plenoptic,zhang2003spectral,Shearlet}, which is intractable to implement within a deep learning pipeline, we derive a novel anti-aliasing framework composed of three sequential operations in the \textit{image domain}, including shearing, downscaling and prefiltering. The proposed framework shows high efficiency in solving the large disparity and non-Lambertian challenges. 

\textbf{Shearing operation.} It translates the disparity range of the input EPI to be centered at zero to minimize the maximum (absolute) disparity value \cite{Tao, ren05}. The shear operation is achieved by shifting each sub-aperture view according to the following equation
\begin{equation}\label{eq:shear}
f_h(E(u,s);\alpha_h)=E(u+s\cdot \alpha_h,s),
\end{equation}
where $f_h$ is the shearing operation and $\alpha_h$ is the shear amount. By shearing the EPI with the optimal rendering disparity $d_{opt}$, the minimum and maximum disparities become $d'_{\min}=\frac{d_{\min}-d_{\max}}{2}$, $d'_{\max}=\frac{d_{\max}-d_{\min}}{2}$, and the optimal rendering disparity $d'_{\textrm{opt}}=0$. Fig. \ref{fig:FA}(c) illustrates the EPI, the Fourier transform and the simplified diagram after the shearing operation.

The shearing operation is a key element of the framework, due to the facts that: 1) it is highly effective for a local EPI patch with large disparity yet small disparity range; 2) it ensures the reconstruction filter to be easily implemented in the image domain, e.g., a Gaussian function. Note that although the shearing may reduce the maximum disparity to a certain extent, the size of the disparity range remains the same, which still induces the aliasing problem.

\textbf{Downscaling operation.} Instead of simply adopting a Gaussian filter for anti-aliasing, we particularly apply the downscaling operation\footnote{The downscaling operation is comparable to the conventional reconstruction filter, whose Fourier analysis is presented in the supplementary.} to increase the sample interval along the spatial dimension~\cite{chai2000plenoptic}, which can be implemented with an anti-aliasing interpolation method~\cite{keys1981cubic} to reduce the spatial resolution. Note that it is a different operation compared with the angular downsampling, we, therefore, use the term ``downscaling'' for distinction. Along with the spatial downscaling, the disparity range is reduced. Therefore, by rescaling a sheared EPI with factor $\alpha_u$, the spectral support will be
\begin{equation}\label{eq:non-lam2}
d'/\alpha_u\Omega_u+\Omega_s=\pm\beta/Z.
\end{equation}
We have $\beta=0$ and $\beta>0$ for Lambertian and non-Lambertian cases, respectively. And the minimum and maximum disparities are $d''_{\min}=\frac{d_{\min}-d_{\max}}{2\alpha_u}$, $d''_{\max}=\frac{d_{\max}-d_{\min}}{2\alpha_u}$.

The downscaling operation forces the original spectrum to shrink towards the $\Omega_u$ axis, separating the original spectrum from its replicas to mitigate the aliasing problem, as shown in Fig. \ref{fig:FA}(d). Afterwards, the width of the reconstruction filter will be increased by a factor of $\alpha_u$, as shown in Fig. \ref{fig:FA}(d) (middle and bottom). One may notice that the pure downscaling operation will cause high-frequency loss according to the downscaling factor. Fortunately, by taking advantage of the deep learning techniques, we can adopt a multi-scale strategy for the balance between aliasing handling and high-frequency maintaining.

\textbf{Prefiltering operation.} The prefiltering operation alleviates the aliasing introduced by the reduced disparity range and the non-Lambertian effect via suppressing the high-frequency components in an EPI~\cite{DBLP}. In the following, we empirically show that the combination of the downscaling and prefiltering is more efficient than using a prefilter merely or the conventional reconstruction filter~\cite{chai2000plenoptic,zhang2003spectral}.

\begin{figure*}
	\begin{center}
		\includegraphics[width=1\linewidth]{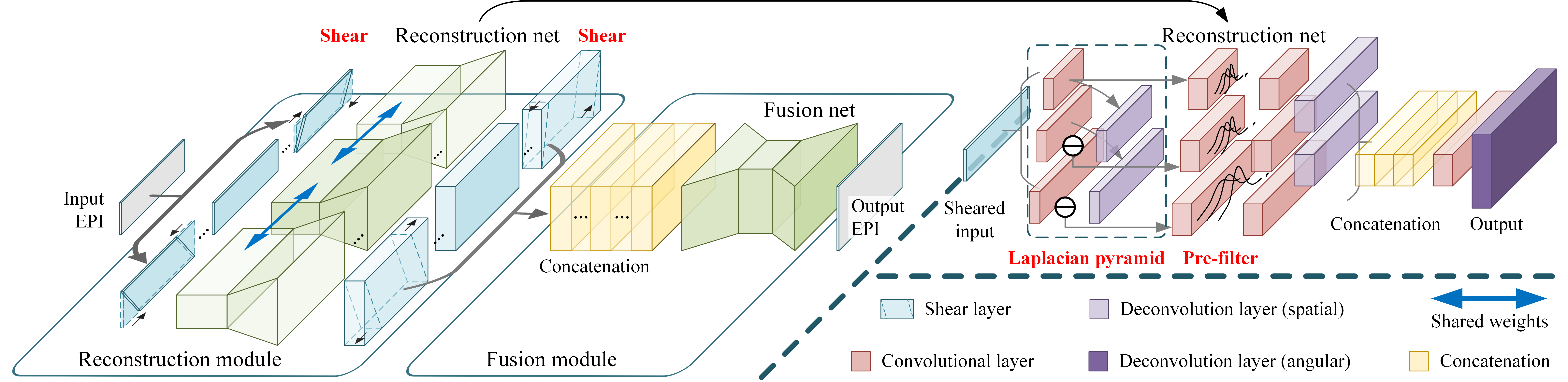}
	\end{center}
	\vspace{-6mm}
	\caption{Architecture of the proposed deep anti-aliasing network (DA$^2$N) based on shearing, downscaling and prefiltering. The DA$^2$N contains a reconstruction module for high angular resolution EPI and a fusion module for blending the EPIs with different shear amounts.}
	\label{fig:CNN}
	\vspace{-3mm}
\end{figure*}

In this paper, we implement the prefiltering using a Gaussian kernel $\kappa$ in image domain, which is equivalent to the reconstruction filtering after taking the shearing with optimal disparity \cite{chai2000plenoptic}. Note that the Gaussian kernel $\kappa$ in image domain can be transformed as $\mathcal{F}_{\kappa}$, a Gaussian function in Fourier domain
\begin{equation}\label{eq:Fourierkernel}
\kappa(u;\sigma)=\frac{1}{\sqrt{2\pi}\sigma}e^{-\frac{u^2}{2\sigma^2}}\Leftrightarrow\mathcal{F}_{\kappa}(\Omega_u;\sigma)=e^{-\frac{\Omega_u^2}{1/(2\pi^2\sigma^2)}},
\end{equation}
where $\sigma$ is the shape parameter. Then the prefiltering or reconstruction filtering in Fourier domain equals to the convolution with kernel in image domain.

The prefilter $\mathcal{F}_{\kappa}$ is designed via the criteria that, the amplitude of a reference aliasing point $P_a$ should not exceed a baseline spectrum amplitude $\gamma$ after being filtered
\begin{equation}\label{eq:decrease}
\mathcal{F}_{E_{LR}}(P_a)\mathcal{F}_{\kappa}(\Omega_{u, P_a};\sigma)\leq\gamma,
\end{equation}
where $\mathcal{F}_{E_{LR}}$ is the Fourier transform of the EPI $E_{LR}$, the reference aliasing $P_a$ has the lowest spatial frequency, $\Omega_{u,P_a}$ is the $\Omega_u$ coordinate of point $P_a$, as shown in Fig. \ref{fig:FA}(c) and (d). We can set different $\gamma$ for different degrees of anti-aliasing. Substitute Eqn. \ref{eq:Fourierkernel} into Eqn. \ref{eq:decrease}, we then have the shape parameter $\sigma$ as
\begin{equation}\label{eq:shape}
\sigma\geq\sqrt{\frac{1}{2\pi^2\cdot\Omega_{u, P_a}^2}\ln\left(\frac{\mathcal{F}_{E_{LR}}(P_a)}{\gamma}\right)},
\end{equation}
where $\Omega_{u, P_a} = \alpha_u[(\Omega_{s, P_a}\pm\beta_{P_a}/Z_{P_a})/(-d_{P_a}')]$ according to Eqn. \ref{eq:non-lam2}. As the depth will not be influenced by the shearing nor the downscaling operation, we consider $\beta_{P_a}/Z_{P_a}$ as a constant. Also, the vertical coordinate $\Omega_{s, P_a}$ and the disparity after the shearing $d_{P_a}'$ are fixed. We then have
\begin{equation}\nonumber
\sigma\propto\frac{1}{\alpha_u}\sqrt{\ln\left(\frac{\mathcal{F}_{E_{LR}}(P_a)}{\gamma}\right)}.
\end{equation}

The $\sigma-\alpha_u$ curves with different baseline amplitude $\gamma$ in range $[5,25]$ are plotted in the bottom of Fig. \ref{fig:FA}(e). On the one hand, the shape parameter $\sigma$ will be smaller with the increasing of the downscaling factor $\alpha_u$ (as shown in the dotted line). On the other hand, the amplitude of point $P_a$ will also be attenuated as the increase of $\alpha_u$. Therefore, the final $\sigma$ of the prefilter will be much smaller after the downscaling operation. In other words, the combination of downscaling and per-filtering is more efficient than simply applying a prefilter or a reconstruction filter on the original signal. The top of Fig. \ref{fig:FA}(e) shows the reconstructed EPI using the introduced framework including shear, downscaling and per-filter operations.


\section{Deep Anti-Aliasing Neural Network}\label{Sec:network}
\begin{figure}
	\begin{center}
		\includegraphics[width=1\linewidth]{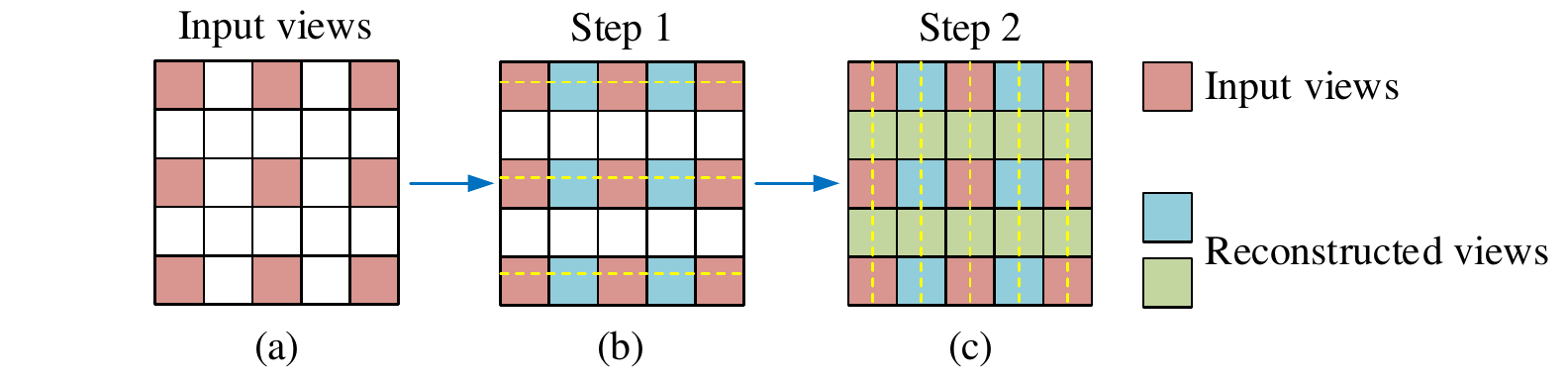}
	\end{center}
	\vspace{-6mm}
	\caption{Two-step strategy for reconstructing a 4D LF from 2D EPIs.}
	\label{fig:EPI2LF}
	\vspace{-3mm}
\end{figure}

In this section, we present DA$^2$N, a Deep Anti-Aliasing Neural Network based on the Fourier analysis on the shearing, downscaling and prefiltering in Sec. \ref{Sec:Problem}. The network contains a reconstruction module for high angular resolution EPI based on the shearing, downscaling and prefiltering, and a fusion module for blending the EPIs (feature maps) with different shear amounts. Fig. \ref{fig:CNN} visualizes the network structure of DA$^2$N, which infers high angular resolution EPIs in an end-to-end manner. For reconstructing a 4D LF from the 2D EPIs, a two-step strategy simplified from the hierarchical reconstruction strategy in~\cite{WuEPICNN2018} is adopted, as illustrated in Fig. \ref{fig:EPI2LF}. EPIs $E_{v^*,t^*}(u,s)$ extracted from the horizontal views (marked as blue in Fig. \ref{fig:EPI2LF}(b)) are applied for reconstructing LF with high resolution in the $s$ dimension. Similarly, EPIs $E_{u^*,s^*}(v,t)$ extracted from the vertical views (marked as green in Fig. \ref{fig:EPI2LF}(c)) are applied for reconstructing LF with high resolution in the $t$ dimension.

\subsection{Reconstruction module}\label{Sec:reconstruction}
The reconstruction module embeds the shearing, downscaling and prefiltering operations within a deep learning framework. The first step of the module is shearing the input low angular resolution EPI $E_{LR}$ with a set of shear amounts $\boldsymbol{H}$, generating multiple streams $f_h(E_{LR};\alpha_h)$ as shown in Fig. \ref{fig:CNN}. Next, the sheared inputs $f_h(E_{LR};\alpha_h)$ are fed into a reconstruction net (Fig. \ref{fig:CNN}, top right) to infer high angular resolution feature maps based on the downscaling and prefiltering. Note that the weights in the reconstruction net are shared for the input with different shear amounts. The final step of the reconstruction module is recovering the high angular resolution feature maps from the former shear operation. We implement this step through another shear operation with the amount $-\alpha_h/\alpha_s$, where $\alpha_s$ is the upsampling factor in the angular dimension.

\begin{figure}
\begin{center}
\includegraphics[width=1\linewidth]{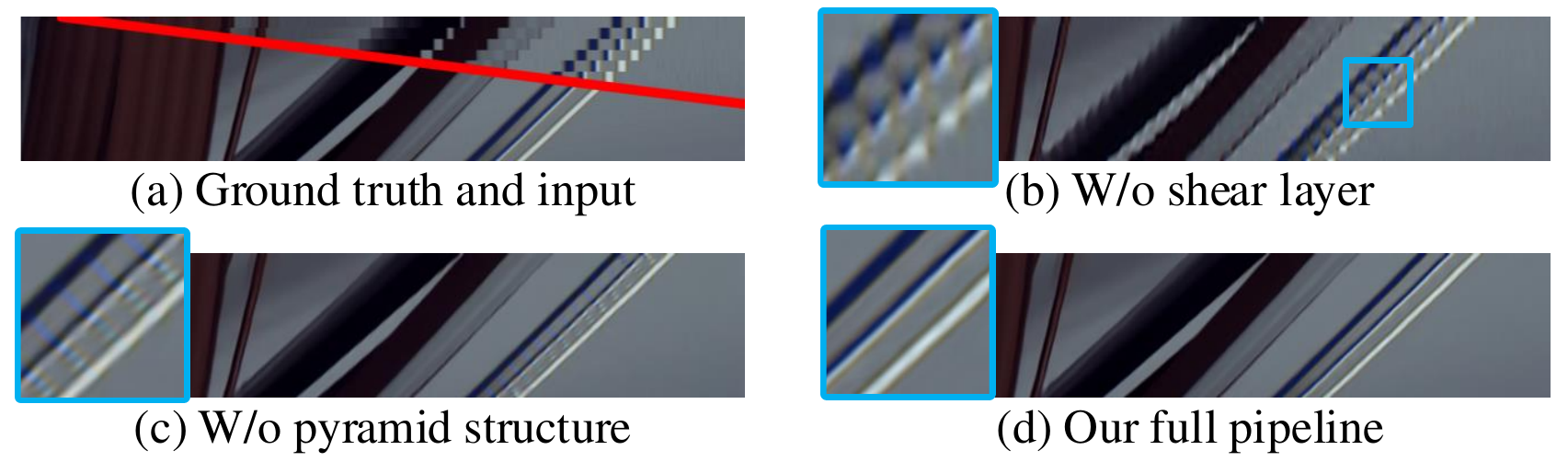}
\end{center}
\vspace{-6mm}
   \caption{The reconstructed LF (or EPI) benefits from the proposed shear layer as well as the downscaling structure. (a) Ground truth and input EPIs; (b) EPI without shear layer; (c) EPI without the learning-based Laplacian pyramid structure; (d) EPI with our proposed pipeline.}
\label{fig:ablation}
\vspace{-3mm}
\end{figure}

To construct a \textbf{shear layer} in the neural network, we extend the operation described in Eqn. \ref{eq:shear} to a general form
\begin{equation}\nonumber
f_h(\phi(u,s,c);\alpha_h)=\phi(u+(s-\frac{S}{2})\cdot \alpha_h,s,c),
\end{equation}
where $S$ is the number of pixels in the angular dimension, $\alpha_h\in\boldsymbol{H}$ is the shear amount and $\phi(u,s,c)$ is a 3D tensor with a spatial dimension $u$, an angular dimension $s$ and a channel dimension $c$. Note that $\phi$ can be the input EPI as well as a feature in the neural network. We use bilinear interpolation for the shear layer described in this paper. And the blank elements in a sheared tensor are set to zero. With the proposed shear layer, the network is able to reconstruct LF with view consistency when encounter with the large disparity challenge, see the comparison demonstrated in Fig. \ref{fig:ablation} (b) and (d). 

A similar utilization of the shear operation in a learning-based method can be found in~\cite{wu2019learning}. However, the proposed pipeline is trained to directly reconstruct a high angular resolution EPI, but the network in~\cite{wu2019learning} is designed to decide whether the EPI is sheared by the optimal rendering disparity. In addition, the proposed pipeline has two fundamental differences in terms of the shear layer as follows: 1) The shear layer in the proposed pipeline is applicable not only to a 2D EPI but also to a feature map in the network, and thus, is embedded into the neural network for end-to-end optimization. In contrast, the shear operation in~\cite{wu2019learning} is only applied for pre- and post-processing of the 2D EPI; 2) We predict multiple high angular resolution EPIs, one for each shear amount, and blend them in the high-level feature domain (please refer to Sec. \ref{Sec:fusion}). While the method in~\cite{wu2019learning} feeds a single sheared EPI into the network, and blends the reconstructed EPIs through weighted average.

In the following, we introduce the reconstruction net that takes the sheared EPI $f_h(E_{LR};\alpha_h)$ as input, where a learning-based Laplacian pyramid is proposed to implement the \textbf{downscaling operation} described in Sec. \ref{Sec:Problem2}. Specifically, we first adopt three convolutional operations with strides $[4,2,1]$ in the spatial dimension (i.e., $\alpha_u=[4,2,1]$) to construct a multi-scale structure; then upsample the first and the second feature layers using deconvolution with stride 2; finally, we construct residual layers by subtracting each upsampled feature from the corresponding layer in the multi-scale structure. Please refer to Fig. \ref{fig:CNN} (top right) for the network connection. The proposed learning-based Laplacian pyramid owns the following superiority: 1) a downscaled feature layer is provided for addressing the aliasing issue echoing with the analysis in Sec. \ref{Sec:Problem2}; 2) a learning-based representation of low-frequency and high-frequency for extracting the aliasing information. As illustrated in Fig. \ref{fig:ablation} (c) and (d), without the proposed pyramid structure, the reconstructed EPI (\ref{fig:ablation} (c)) shows obvious aliasing artifacts.

For the \textbf{prefilter layer}, we use 1D convolutional operations with different kernel sizes for different spatial scales. We use the Gaussian function to initialize the trainable weights. Moreover, we adopt multiple shape parameters in each layer to achieve an effect of different degrees of anti-aliasing, which enables the network for addressing different degrees of aliasing. Specifically, we extend the Gaussian kernel described in Eqn. \ref{eq:Fourierkernel} to a higher dimension as follows
\begin{equation}\nonumber
\theta_{p}(u,c)=\frac{\exp({-\frac{u^2}{2\sigma^2(c)}})}{\sum\limits_{u}{\exp({-\frac{u^2}{2\sigma^2(c)}}})},
\end{equation}
where $\theta_{p}$ is the 1D convolutional kernel in the prefilter layer and $\sigma(c)$ is the shape parameter for channel $c$. For each spatial scale in the prefilter layer, the maximum value of the shape parameter $\sigma_{\max}$ is selected according to the theoretical analysis in Sec. \ref{Sec:Problem}, and $\sigma(c)$ for each channel $c$ is uniformly distributed from zero to $\sigma_{\max}$. To be consistent with the definition of filtering, bias and activation function are not applied in the prefilter layer\footnote{After the end-to-end training, the filters in the prefilter layers are no longer simple Gaussian functions but are still some kinds of low-pass filters. Please refer to the supplementary for more details.}.

\begin{table}
\caption{Architecture of the reconstruction net, where \textbf{k} is the kernel size (spatial/angular), \textbf{str} is the stride (spatial/angular), \textbf{chn} is the number of channels (input/output), and \textbf{Input} is the input layer with ``$\ominus$'' denoting element-wise subtraction and ``;'' denoting concatenation in the channel dimension.}
\vspace{-5mm}
\begin{center}
\begin{tabular}{p{1.5cm}<{\centering}p{0.8cm}<{\centering}p{0.9cm}<{\centering}p{0.8cm}<{\centering}p{2.6cm}<{\centering}}
\toprule
\textbf{Layer} & \textbf{k} & \textbf{str} & \textbf{chn} & \textbf{Input} \\
\hline
conv1\_1 & [5, 5] & [4, 1] & 1/10 & $f_h(E_{LR})$\\
conv1\_2 & [5, 5] & [2, 1] & 1/10 & $f_h(E_{LR})$ \\
conv1\_3 & [5, 5] & [1, 1] & 1/10 & $f_h(E_{LR})$ \\
deconv2\_1 & [3, 3] & [2, 1] & 10/10 & conv1\_1\\
deconv2\_2 & [5, 5] & [2, 1] & 10/10 & conv1\_2\\
conv2\_1 & [5, 1] & [1, 1] & 10/20 & conv1\_1\\
conv2\_2 & [11, 1] & [1, 1] & 10/20 & conv1\_2$\ominus$deconv2\_1\\
conv2\_3 & [21, 1] & [1, 1] & 10/20 & conv1\_3$\ominus$deconv2\_2\\
conv3\_1 & [3, 3] & [1, 1] & 20/27 & conv2\_1\\
conv3\_2 & [3, 3] & [1, 1] & 20/27 & conv2\_2\\
conv3\_3 & [3, 3] & [1, 1] & 20/27 & conv2\_3\\
deconv4\_1 & [5, 5] & [4, 1] & 27/27 & conv3\_1\\
deconv4\_2 & [5, 5] & [2, 1] & 27/27 & conv3\_2\\
\multirow{2}*{conv4} & \multirow{2}*{[3, 3]} & \multirow{2}*{[1, 1]} &  \multirow{2}*{81/81} & deconv4\_1;\\
& & & & deconv4\_2; conv3\_3\\
deconv5 & [9, 9] & [1, $\alpha_s$] & 81/27 & conv4\\
\bottomrule
\end{tabular}
\end{center}
\label{table:reconstruction}
\vspace{-4mm}
\end{table}

Straight after the prefilter layer, we apply a convolutional layer for both the spatial and the angular dimensions, and a deconvolution layer (spatial) for upscaling the feature maps to the original spatial resolution. Then the three streams are concatenated together in the channel dimension, being followed by a convolutional layer with batch normalization \cite{ioffe2015batch}. The last layer is a deconvolution layer (angular) that is designed to reconstruct high angular resolution EPI (feature maps) through upsampling. Each layer in the reconstruction net except for the prefilter layer and the angular deconvolution layer is followed by a leaky ReLU. Please refer to Table. \ref{table:reconstruction} for detailed specification of the reconstruction net.

In summary, the entire reconstruction module $f_R$ can be described as follows
\begin{equation}\label{eq:reconstruct}
f_R(E_{LR};\theta_r)=\{f_h(f_r(f_h(E_{LR};\alpha_h^{(i)});\theta_r);-\frac{\alpha_h^{(i)}}{\alpha_s})\},
\end{equation}
where $\alpha_h^{(i)}$ is the $i$-th shear amount and $f_r(\cdot;\theta_r)$ denotes the reconstruction net with trainable parameters $\theta_r$. The shear amount $\alpha_h$ and angular upsampling factor $\alpha_s$ are two hyperparameters, set as $\alpha_h\in[-9, -6, -3, 0, 3, 6, 9]$ and $\alpha_s=3, 4$. Note that the shear values can be readjusted without retraining, and the reconstruction scale is flexible via cascade.

\subsection{Fusion module}\label{Sec:fusion}
The fusion module serves to blend multiple feature layers with different shear amounts, namely different candidates of the reconstructed EPI, as shown in Fig. \ref{fig:CNN}. Note that only the candidate sheared by the optimal or close to the optimal rendering disparity will get a high quality reconstruction result. In this way, the overall framework is able to select optimal rendering disparity without explicit disparity estimation.

The input of the fusion module is a 3D tensor, i.e., a high angular resolution feature map, rather than a 2D EPI slice, to prevent unnecessary feature extractions from 2D EPIs. We first concatenate the sheared layers yielded by the reconstruction module along the channel dimension. The last step contains a fusion net that produces the final high angular resolution EPI by blending the multiple sheared feature layers together. The proposed fusion net follows the U-net structure~\cite{alperovich2018light}, while it varies in terms of: 1) a convolutional layer with kernel size $[1,1]$ is set in the front of the fusion net to compress the feature maps; 2) the encoder (decoder) downscales (upscales) the feature maps only in the spatial dimension. Each layer except for the last layer is followed by a leaky ReLU. Table \ref{table:fusion} lists the detailed architecture of the fusion net as well as the previous concatenation layer.

\begin{table}
\caption{Architecture of the fusion net.}
\vspace{-5mm}
\begin{center}
\begin{tabular}{p{1.5cm}<{\centering}p{0.8cm}<{\centering}p{0.8cm}<{\centering}p{0.8cm}<{\centering}p{2.6cm}<{\centering}}
\toprule
\textbf{Layer} & \textbf{k} & \textbf{str} & \textbf{chn} & \textbf{Input} \\
\hline
conv1\_1 &[1, 1]&[1, 1]&189/27 & $f_h^{(1)}(f_r);\cdots;f_h^{(7)}(f_r)$\\
conv1\_2 & [3, 3] & [2, 1] & 27/54 & conv1\_1 \\
conv2\_1 & [3, 3] & [1, 1] & 54/54 & conv1\_2\\
conv2\_2 & [3, 3] & [2, 1] & 54/54 & conv2\_1\\
conv3\_1 & [3, 3] & [1, 1] & 54/54 & conv2\_2\\
conv4 & [3, 3] & [1, 1] & 54/54 & conv3\_1\\
deconv5\_1 & [5, 5] & [2, 1] & 54/54 & conv4\\
conv5\_2 & [1, 1] & [1, 1] & 108/54 & deconv5\_1; conv2\_1\\
deconv6\_1 & [5, 5] & [2, 1] & 54/27 & conv5\_2\\
conv6\_2 & [1, 1] & [1, 1] & 54/27 & deconv6\_1; conv1\_1\\
conv7 & [9, 9] & [1, 1] & 27/1 & conv6\_2\\
\bottomrule
\end{tabular}
\end{center}
\label{table:fusion}
\vspace{-4mm}
\end{table}

Combining with the reconstruction module described in Eqn. \ref{eq:reconstruct}, the proposed DA$^2$N with all the trainable parameters can be represented as
\begin{equation}\label{eq:network}
f_{D}(E_{LR};\boldsymbol{\theta})=f_F(f_R(E_{LR};\theta_r);\theta_f),
\end{equation}
where $\boldsymbol{\theta}=\{\theta_r, \theta_f\}$, and $f_{F}(\cdot;\theta_f)$ is the fusion net with trainable parameters $\theta_f$.

\section{Network Training}
\subsection{Training dataset}\label{Sec:dataset}
A sufficiently large set of LFs with plenty views for extracting large-disparity LFs or with non-Lambertian property is favored for the training procedure. Unfortunately, commonly-used LF dataset (LFs from HCI datasets~\cite{HCI} or a Lytro Illum~\cite{Lytro}) is lacking data with non-Lambertian effect. Meanwhile, existing LF datasets suffer small disparity range and limited angular resolution. For example, in the HCI datasets, the disparity range is around $\pm2$ pixel and angular resolution is $9\times9$. For the Lytro LF, the disparity is nearly within 1 pixel range, and the valid views are $8\times8$. With an angular downsampling factor of 3, these datasets only provide an input with around 6 and 3 disparity ranges, respectively. 
\begin{figure}
\begin{center}
\includegraphics[width=1\linewidth]{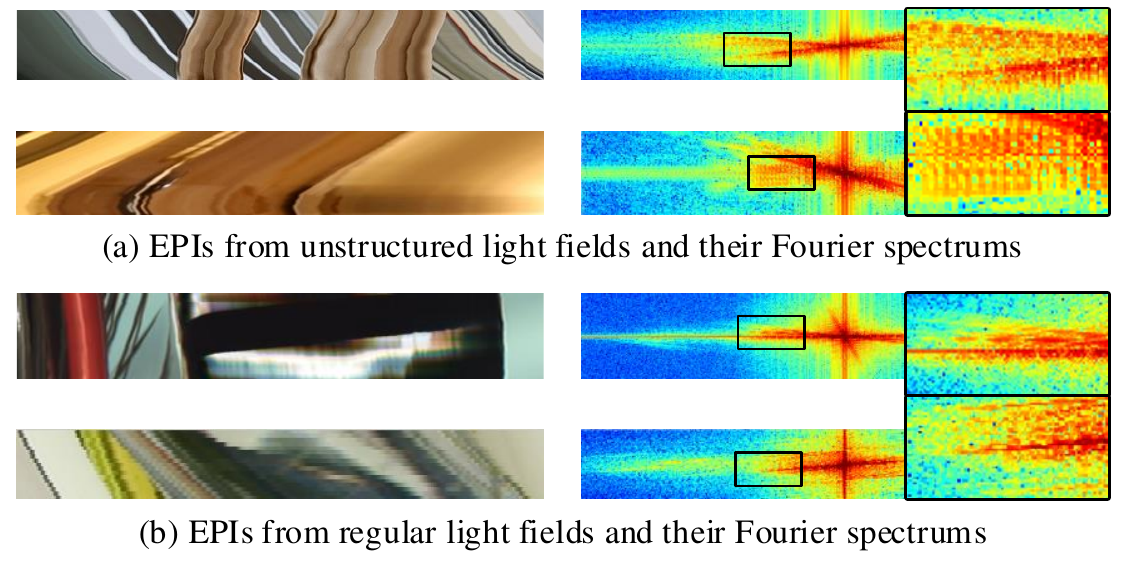}
\end{center}
\vspace{-6mm}
   \caption{Pseudo EPIs and regular EPIs in non-Lambertian scenes can be described by curves and variation of pixel intensity (color). These two EPIs show the same property in the Fourier domain as well.}
\label{fig:structure}
\vspace{-3mm}
\end{figure}

Considering the above problems, we adopt EPIs from regular (calibrated) LFs for training from scratch and \textbf{pseudo EPIs from unstructured LFs} for fine-tuning. Our key insight is that pseudo EPIs exactly accord with the non-Lambertian property in regular EPIs. In the following, we explain this insight by expanding the Fourier analysis in Sec. \ref{Sec:Problem1} to the pseudo EPIs from unstructured LFs.

An unstructured LF may be generated from an irregularly moving camera, including a non-linear movement along the main angular dimension\footnote{In this paper, we only consider a 3D unstructured LF, namely, the LF is recorded along a main angular dimension.}, a slight shaking along the other angular dimension or around the normal of the camera plane. The former one introduces ``curves'' in the recorded EPI, as shown in the top of Fig. \ref{fig:structure}(a). While the latter one leads to a sporadic ``appearance'' of certain pixels along the angular dimension, i.e., abrupt variation of intensity (or color), as shown in the bottom of Fig. \ref{fig:structure}(a). In the first case, the curve in a pseudo EPI can be split into lines with a series of different disparities. Therefore, its spectrum will be $D\Omega_u+\Omega_s=0$, where $D$ is the set of disparities from the curve (as shown in the top right spectrum of Fig. \ref{fig:structure}(a)). In the second case, the intensity (color) variation can be approximated by a band-limited signal \cite{zhang2003spectral}, thus, its spectrum is $d\Omega_u+\Omega_s=\pm\beta'/Z$, where $\beta'$ is the band width of the signal. By combining these two case, we then have the Fourier spectrum described as $D\Omega_u+\Omega_s=\pm\beta'/Z$ (see the bottom right spectrum of Fig. \ref{fig:structure}(a)). In a non-Lambertian scene, the regular EPI can be also regarded as the combination of curves and variation of pixel intensity, which can be clearly seen in Fig. \ref{fig:structure}(b). And the spectrum of the EPI also conforms to the formula $D\Omega_u+\Omega_s=\pm\beta'/Z$ (as shown in the right spectra of Fig. \ref{fig:structure}(b)).

In addition, unstructured LFs can easily provide a high angular resolution (up to $600$ pixels in our dataset) due to the acquisition method (e.g., by a hand-hold video camera), therefore, is able to offer EPIs with a large disparity. In conclusion, the pseudo EPI dataset provides the same non-Lambertian description and a better data abundance for the network training.

\subsection{Objective}
Based on the high angular resolution EPI reconstruction pipeline defined in Eqn. \ref{eq:network}, the objective is to minimize a loss function $\mathcal{L}(\cdot)$ between the high angular resolution EPI label $E_{HR}$ and the reconstructed EPI $f_{D}(E_{LR};\boldsymbol{\theta})$ as
\begin{equation}\nonumber
\min\limits_{\boldsymbol{\theta}}\sum\limits_{\langle E_{LR},E_{HR}\rangle}{ \mathcal{L}(f_{D}(E_{LR};\boldsymbol{\theta}),E_{HR})},
\end{equation}
where $\langle E_{LR},E_{HR}\rangle$ is the training set of low and high angular resolution EPI tuples.

We combine two terms in the loss function $\mathcal{L}$, a $L_1$ distance and a perceptual loss (also known as high-level feature matching loss) between the label EPI and the reconstructed EPI. The perceptual loss \cite{dosovitskiy2016generating,johnson2016perceptual} can be computed from part of the feature layers in the autologous network \cite{dosovitskiy2016generating} or other pre-trained networks \cite{johnson2016perceptual}, e.g., the commonly-used VGG-16 \cite{simonyan2014very}. To prevent loss fluctuation in the training procedure, we adopt a pre-trained network that is specifically trained on LF dataset. The network is an encoder-decoder with a similar structure of the fusion net. While the main differences are that no skip connections are used and the downscales (upscales) are performed in both the spatial and angular dimensions. We apply the same training dataset as introduced in Sec. \ref{Sec:dataset} for the encoder-decoder. The loss function is then defined as
\begin{equation}\label{eq:loss}
\begin{split}
\mathcal{L}(\hat{E}_{HR},E_{HR})&=\Vert\hat{E}_{HR}-E_{HR}\Vert_1\\
&+\sum\limits_l\lambda_{feat}^{(l)}\Vert\phi^{(l)}(\hat{E}_{HR})-\phi^{(l)}(E_{HR})\Vert_1,
\end{split}
\end{equation}
where $\{\phi^{(l)}\}$ is a set of layers in the encoder including conv1\_1, conv2\_1 and conv3\_1, and $\lambda_{feat}=[0.2, 0.2, 0.1]$ is a set of hyperparameters for the perceptual loss.

\begin{figure*}
\begin{center}
\includegraphics[width=1\linewidth]{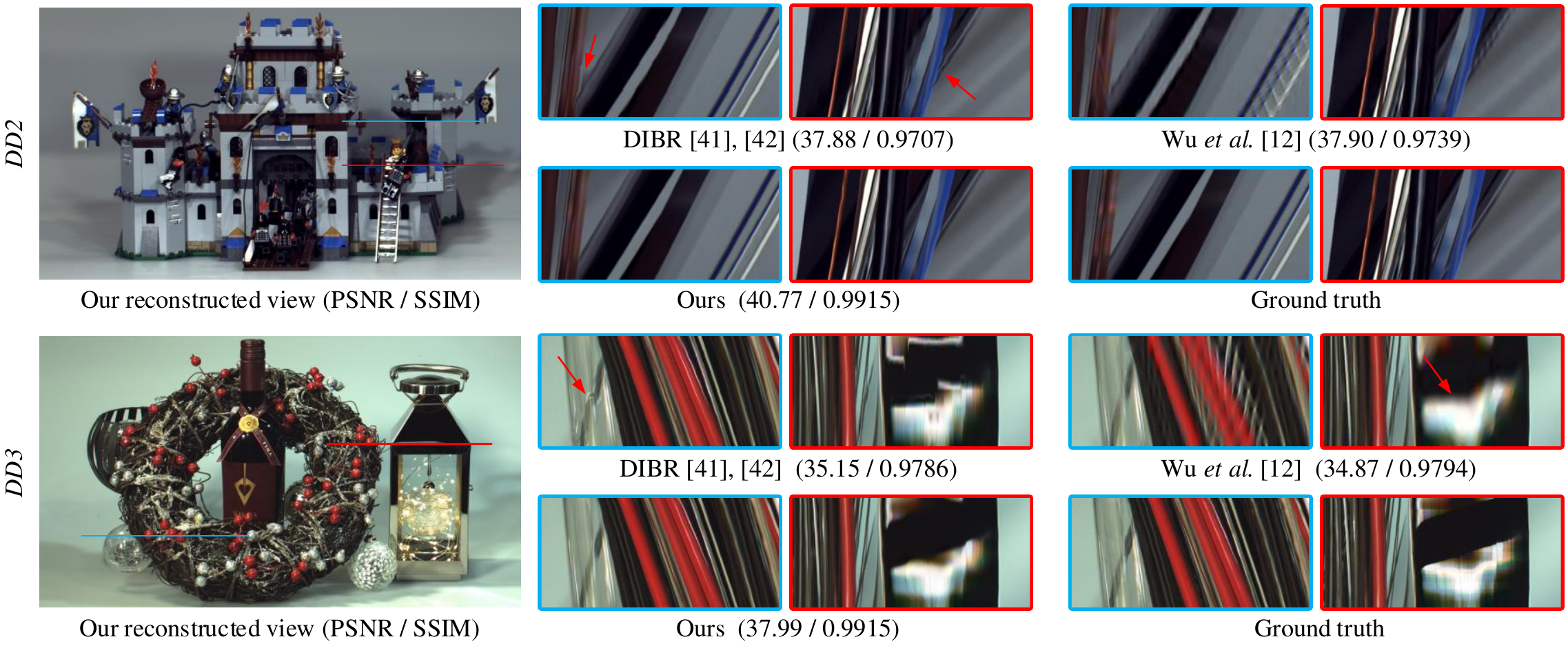}
	\vspace{-10mm}
\end{center}
   \caption{Comparison of the results ($\times16$ upsampling) on the LFs from the ICME DSLF dataset~\cite{ICME2018}. The results show one of our reconstructed view, EPIs extracted from LFs reconstructed by each methods. The PSNR and SSIM values are averaged on the reconstructed views (input views are excluded).}
\label{fig:Result1}
\end{figure*}

\begin{table*}
\caption{Quantitative results (PSNR/SSIM) of reconstructed LFs on the LFs from the ICME DSLF dataset~\cite{ICME2018}.}
\vspace{-8mm}
\begin{center}
\begin{tabular}{clccccccc}
\toprule
& \multirow{2}*{Method} & \multicolumn{2}{c}{\textit{DD1}} & \multicolumn{2}{c}{\textit{DD2}} & \multicolumn{2}{c}{\textit{DD3}} & \multirow{2}*{Average}\\
\cmidrule(r){3-4}\cmidrule(r){5-6}\cmidrule(r){7-8}
& & (7px) $\times8$ & (14px) $\times16$ & (7px) $\times8$ & (14px) $\times16$ & (7px) $\times8$  & (14px) $\times16$\\
\hline
\multirow{5}*{Baseline} & DIBR~\cite{Acc,CDSD13}& 39.93/0.9884 & 39.31/0.9876 & 37.99/0.9743 & 37.88/0.9707 & 36.09/0.9860 & 35.15/0.9786 & 37.73/0.9809\\
& Kalantari~\textit{et al.}~\cite{DoubleCNN}& 45.66/0.9901 & 42.97/0.9840 & 38.95/0.9712 & 35.94/0.9631 & 35.88/0.9741 & 31.92/0.9593 & 38.55/0.9736\\
& Wu~\textit{et al.}~\cite{WuEPICNN2018}& 44.81/0.9905 & 42.31/0.9761 & 40.58/0.9833 & 37.90/0.9739 & 37.83/0.9931 & 34.87/0.9794 & 39.72/0.9827\\
& Yeung~\textit{et al.}~\cite{YeungECCV2018}& 47.10/0.9965 & 41.36/0.9836 & 39.61/0.9915 & 35.45/0.9834 & 39.96/0.9951 & 32.83/0.9749 & 39.39/0.9875\\
& Wu~\textit{et al.}~\cite{wu2019learning}& 46.89/0.9899 & 44.81/0.9860  & 40.87/0.9716  & 38.65/0.9668  & 37.96/0.9859  & 35.83/0.9771 & 40.84/0.9796\\
\hline
\multirow{5}*{Ablation} & w/o shear &46.78/0.9948 &44.05/0.9930 & 41.55/0.9922 & 39.32/0.9903 & 40.08/0.9951 & 36.68/0.9903 &41.41/0.9926\\
& w/o downscale &46.71/0.9964 &44.49/0.9948 & 41.38/0.9921 & 39.22/0.9893 & 39.79/0.9942 & 36.41/0.9890 & 41.33/0.9926\\
& w/o prefilter &47.92/0.9972 &45.87/0.9949 & 42.42/0.9923 & 40.52/0.9912 & 41.08/0.9953 & 37.51/0.9897 & 42.55/0.9934\\
& w/o pseudo EPIs & 46.53/0.9962 & 41.21/0.9834 & 41.93/0.9920 & 38.14/0.9862 & 40.58/0.9954 & 33.60/0.9781 & 40.33/0.9886\\
& Our proposed &\textbf{48.23}/\textbf{0.9971} &\textbf{46.19}/\textbf{0.9956} & \textbf{42.65}/\textbf{0.9934} & \textbf{40.77}/\textbf{0.9915} & \textbf{41.38}/\textbf{0.9960} & \textbf{37.99}/\textbf{0.9915} & \textbf{42.87}/\textbf{0.9942}\\
\bottomrule
\end{tabular}
\begin{tablenotes}
\textit{Ablation studies are performed for the network without the shear layers (``w/o shear'' for short), without the Laplacian pyramid structure for downscaling (``w/o downscale'' for short), without prefilter layers (``w/o prefilter'' for short) and without pseudo EPIs from unstructured LFs for fine-tuning (``w/o pseudo EPIs'' for short). The disparity range ($d_{\max}-d_{\min}$) between the neighboring views in the input light field is listed in the table.}
\end{tablenotes}
\end{center}
\label{table:Result1}
\end{table*}

\subsection{Implementation details}
The upsampling scale $\alpha_s$ in the angular dimension is fixed for the training, but the reconstruction scale is flexible by combining model cascade and bicubic downsampling (angular). For instance, for interpolation $\times7$, we will first achieve a $\times9$ reconstruction with two cascades ($\alpha_s=3$) and then downsample the resulting LF to the desired angular resolution. The training is performed on the Y channel (i.e., the luminance channel) of the YCbCr color space. To speed up the training procedure, we apply patch-wise training strategy by sampling sub-EPI pairs in the training data, and then applying mini-batches of size 28. We use zero padding in each layers according to the filter size.

Instead of training the DA$^2$N using mixed datasets from the regular LFs and the unstructured LFs, we divide the training into the following two steps.

\textit{Training from scratch on regular LFs.} In this step, about 13K EPIs are extracted from 11 LFs from the Stanford Light Field Archive \cite{StanfordLFdatasets} with stride $20$ in the spatial dimensions. In our implementation, we extract sub-EPIs of size $6\times72$ for the input examples and $16\times72$ for the labels from the EPIs with stride $16$ in the spatial dimensions, and donwnsample the angular dimension using nearest sampling. Through this way, we obtain about 725K examples from the structured LFs.
We initialize the weights of both convolution and deconvolution layers by drawing randomly from a Gaussian distribution with a zero mean and standard deviation $1\times10^{-3}$, and the biases by zero. The prefilter layers are initialized with Gaussion function as mentioned in Sec. \ref{Sec:reconstruction}. We use ADAM solver~\cite{Kingma2014Adam} as the optimization method with learning rate of $1\times10^{-5}$ for the prefilter layers and $4\times10^{-5}$ for the rest of the layers, and $\beta_1=0.9$, $\beta_2=0.999$. The network converges after 600K steps of backpropagation.

\textit{Fine-tuning on unstructured LFs.} In this step, we extract 1888 pseudo EPIs (from 11 unstructured LFs by Y\"{u}cer~\textit{et al.}~\cite{Yucer16}) at $600\times1280$ (or 1920) (angular$\times$spatial) resolution with stride $31/400$ along dimension $v/t$. About 1400K sub-EPIs with resolution $31\times72$ (for the labels) are extracted from pseudo EPIs (stride $92/23$ for the angular/spatial dimension). The learning rate is set to $1\times10^{-6}$ for the prefilter layers and $1\times10^{-5}$ for the rest of the layers. The network converges after 800K steps of backpropagation. The training model is implemented using the \emph{Tensorflow} framework~\cite{TensorFlow}. The entire training steps (training from scratch and fine-tuning) takes about 50 hours on a NVIDIA TITAN Xp.

\begin{figure*}
	\begin{center}
		\includegraphics[width=1\linewidth]{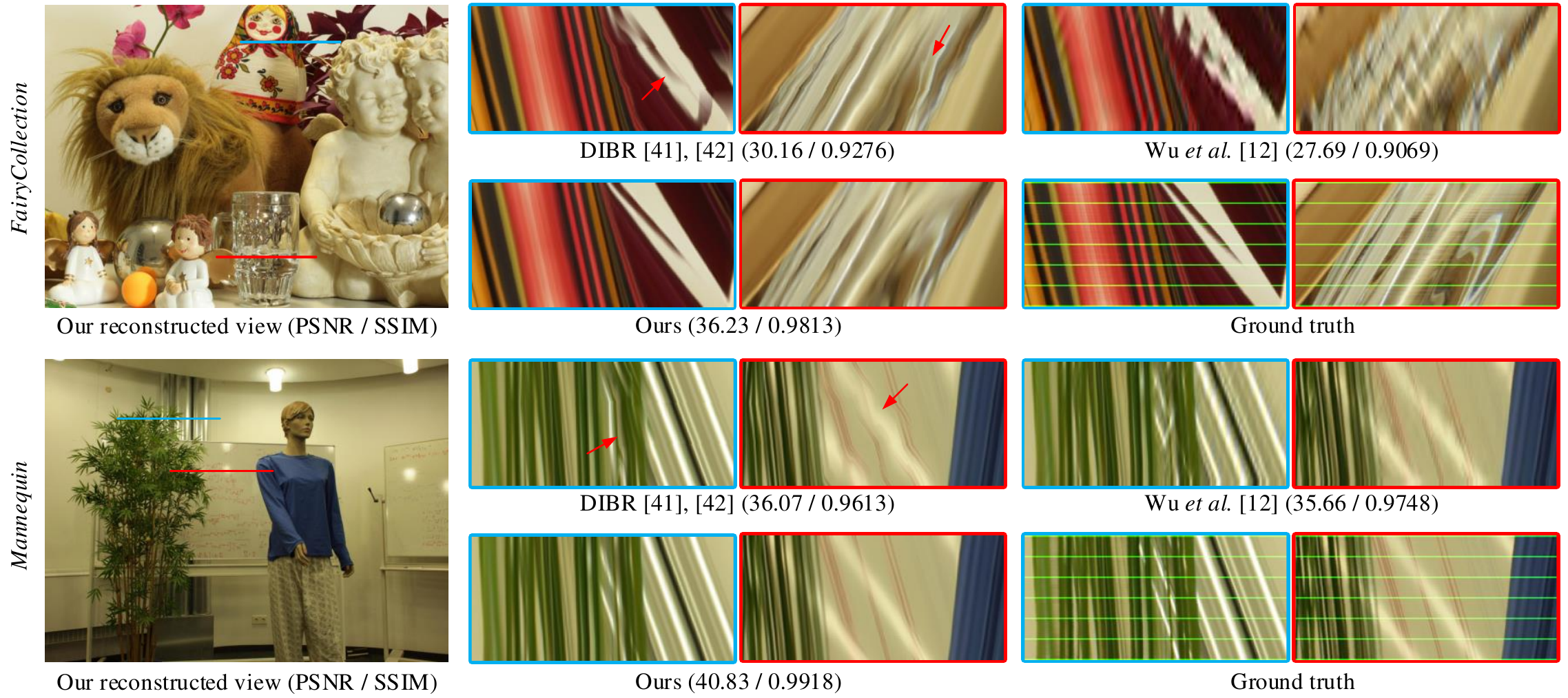}
	\vspace{-10mm}
	\end{center}
	\caption{Comparison of the results ($\times16$ upsampling) on the LFs from the MPI Light Field Archive~\cite{kiran2017towards}. The green lines in the ground truth EPI are input views. The PSNR and SSIM values are averaged on the reconstructed views (input views are excluded).}
	\label{fig:Result1_2}
\end{figure*}

\begin{table*}
\caption{Quantitative results (PSNR/SSIM) of reconstructed LFs ($\times16$) on the LFs from the MPI Light Field Archive~\cite{kiran2017towards}.}\label{tab:result1_2}
\vspace{-3mm}
\begin{center}
\begin{tabular}{lcccccc}
\toprule
Method& \textit{Bikes} (23.5px) & \textit{FairyCollection} (33.5px) & \textit{LivingRoom} (14px) & \textit{Mannequin} (14.5px) & \textit{WorkShop} (23px) & Average\\
\hline
DIBR~\cite{Acc,CDSD13}&  30.03/0.9525 & 30.16/0.9276 & 44.07/0.9860 & 36.07/0.9613 & 32.33/0.9690 & 34.53/0.9593\\
Kalantari~\textit{et al.}~\cite{DoubleCNN}& 27.89/0.9101 & 28.37/0.9231 & 39.95/0.9901 & 36.12/0.9689 & 35.72/0.9677 & 33.61/0.9520\\
Wu~\textit{et al.}~\cite{WuEPICNN2018} & 27.23/0.9045 & 27.69/0.9069 & 40.87/0.9909 & 35.66/0.9748 & 30.08/0.9371 & 32.31/0.9428\\
Yeung~\textit{et al.}~\cite{YeungECCV2018} & 30.16/0.9365 & 30.37/0.9427 & 43.10/0.9896 & 36.74/0.9792 & 33.81/0.9561 & 34.84/0.9608\\
Wu~\textit{et al.}~\cite{wu2019learning} & 31.58/0.9425 & 35.02/0.9721 & 44.56/0.9885 & 38.21/0.9760 &36.47/0.9796 & 37.17/0.9717\\
Our proposed & \textbf{35.79}/\textbf{0.9844} & \textbf{36.23}/\textbf{0.9813} & \textbf{45.91}/\textbf{0.9963} & \textbf{40.83}/\textbf{0.9918}  & \textbf{40.11}/\textbf{0.9936} & \textbf{39.77}/\textbf{0.9895}\\
\bottomrule
\end{tabular}
\begin{tablenotes}
\textit{The disparity range ($d_{\max}-d_{\min}$) between the neighboring views in the input light field is listed in the table.}
\end{tablenotes}
\vspace{-3mm}
\end{center}
\end{table*}

\section{Evaluation and Application}
In this section, we evaluate the proposed DA$^2$N on several datasets, including LFs from gantry system, LFs from a hand-held plenoptic camera (Lytro Illum~\cite{Lytro}) and LFs from camera array system. Note that all these datasets are not overlap with our training data in terms of acquisition geometry. The term ``large disparity'' is defined as the maximum disparity between neighboring views being larger than 5 pixels or the disparity range being more than 10 pixels.

We mainly compare our approach with four state-of-the-art methods, including two depth and learning-based methods by Kalantari~\textit{et al.}~\cite{DoubleCNN} and Wu~\textit{et al.}~\cite{wu2019learning}, and two depth independent and learning-based methods by Yeung~\textit{et al.}~\cite{YeungECCV2018} and Wu~\textit{et al.}~\cite{WuEPICNN2018}. In addition, we perform a series of ablation studies of our approach. The quantitative evaluations are reported by measuring the average PSNR and SSIM~\cite{SSIM} values over the synthesized views (in both figures and tables). Please also refer to the submitted supplementary video for more qualitative results. The code will be available at \url{https://github.com/GaochangWu/DA2N}.

\subsection{Light fields from gantry system}
\label{Sec:Experiment1}
We use the dataset from the ICME 2018 Grand Challenge on Densely Sampled Light Field Reconstruction~\cite{ICME2018} (``ICME DSLF dataset'' for short) and the MPI Light Field Archive~\cite{kiran2017towards}, for the evaluation of LFs from gantry system. In this experiment, we reconstruct a $1\times193$ LF using 13 views for the ICME DSLF dataset~\cite{ICME2018} and a $1\times97$ LF using 7 views for the MPI Light Field Archive~\cite{kiran2017towards}, i.e., $\times16$ upsampling. 

In this experiment, the baseline methods are a conventional depth image-based rendering (DIBR) approach (the approach by Jeon~\textit{et al.}~\cite{Acc} for depth estimation and the approach by Chaurasia~\textit{et al.}~\cite{CDSD13} for view synthesis), the depth independent and learning-based approach by Wu~\textit{et al.}~\cite{WuEPICNN2018} and Yeung~\textit{et al.}~\cite{YeungECCV2018}, and the depth and learning-based approach by Kalentary~\textit{et al.}~\cite{DoubleCNN} and Wu~\textit{et al.}~\cite{wu2019learning}. The baseline methods~\cite{DoubleCNN,YeungECCV2018} are retrained for better comparison.

\begin{figure*}
	\begin{center}
		\includegraphics[width=1\linewidth]{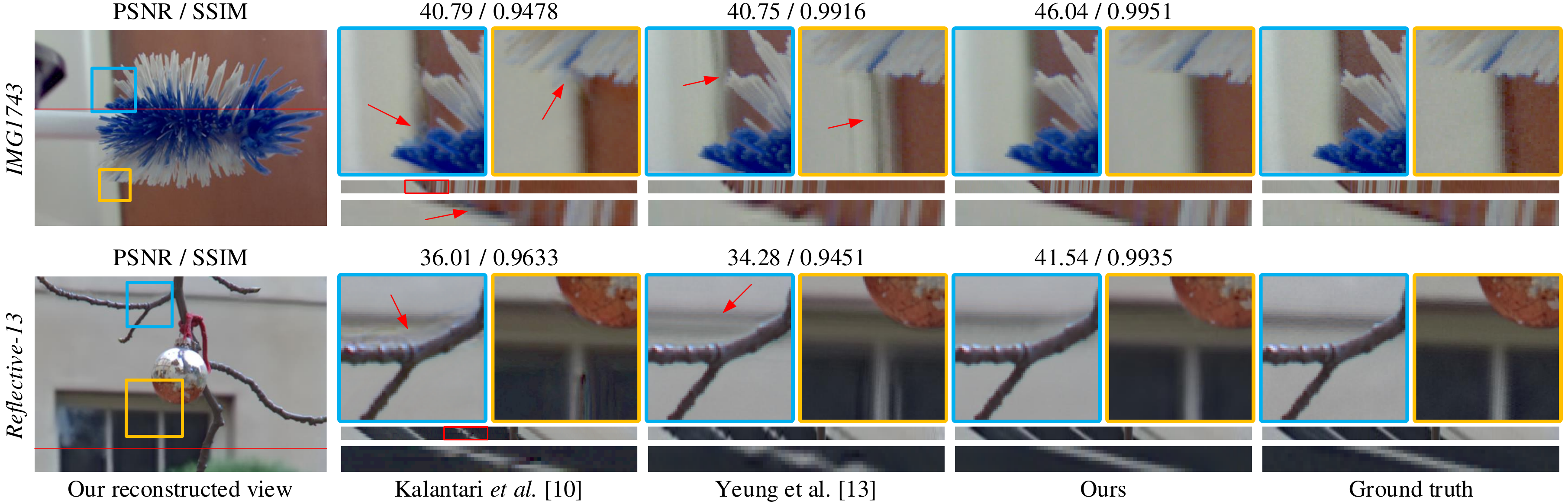}
	\end{center}
	\vspace{-6mm}
	\caption{Comparison of the results on the LFs from Lytro Illum ($\times3$ upsamping). The results show one of our reconstructed views, close-up versions of the image portions in the yellow and blue boxes, and the EPIs located at the red line shown in the left images. We also show a close-up of the portion of the EPIs in the red box. LFs are from the \textit{30 Scenes} set~\cite{DoubleCNN} and the \textit{Refractive and Reflective Surfaces} category~\cite{StanfordLytro}.}
	\label{fig:Result2}
\end{figure*}

\begin{table*}
\caption{Quantitative results (PSNR/SSIM) of reconstructed views on the LFs from plenoptic cameras. }
\vspace{-5mm}
\begin{center}
\begin{tabular}{lp{1.73cm}<{\centering}p{1.73cm}<{\centering}p{1.73cm}<{\centering}p{1.73cm}<{\centering}p{1.73cm}<{\centering}p{1.73cm}<{\centering}p{1.73cm}<{\centering}}
\toprule
\multirow{2}*{Method} & \multicolumn{2}{c}{\textit{30 Scenes}} & \multicolumn{2}{c}{\textit{Reflective \& Refractive Surfaces}} &  \multicolumn{2}{c}{\textit{Occlusions}} & \multirow{2}*{Average}\\
\cmidrule(r){2-3}\cmidrule(r){4-5}\cmidrule(r){6-7}
& $\times3$ & $\times7$ & $\times3$ & $\times7$ & $\times3$ & $\times7$\\
\hline
Kalentary~\textit{et al.}~\cite{DoubleCNN}& 39.62/0.9782 & 38.21/0.9736 & 37.78/0.9713 & 35.84/0.9416 & 34.02/0.9554 & 31.81/0.8945 & 35.61/0.9474\\
Wu~\textit{et al.}~\cite{WuEPICNN2018}& 41.02/0.9875 &36.28/0.9648 & 41.71/0.9886 & 36.48/0.9619 & 38.30/0.9701 & 32.36/0.9240& 37.28/0.9628\\
Yeung~\textit{et al.}~\cite{YeungECCV2018}& \textbf{44.53}/0.9900 & \textbf{39.22}/0.9773 & 42.56/0.9745 & 36.47/0.9472 & 39.27/0.9450 & 32.68/0.9061 & 38.54/0.9510\\
Wu~\textit{et al.}~\cite{wu2019learning}& 42.46/0.9861 & 38.21/0.9711 & 42.28/0.9773 & 36.55/0.9534 & 38.11/0.9443 & 32.19/0.9067& 37.73/0.9509\\
Our proposed & 43.69/\textbf{0.9949} & 38.99/\textbf{0.9863} & \textbf{43.25}/\textbf{0.9906} & \textbf{36.72}/\textbf{0.9751} & \textbf{39.82}/\textbf{0.9706} & \textbf{33.14}/\textbf{0.9504} & \textbf{38.76}/\textbf{0.9748}\\
\bottomrule
\end{tabular}
\begin{tablenotes}
\textit{The metrics are averaged over 113 LFs.}
\end{tablenotes}
\end{center}
\label{table:Result2}
\vspace{-4mm}
\end{table*}

Fig. \ref{fig:Result1} shows the results ($\times16$ upsampling) of the case \textit{DD2} and \textit{DD3} from the ICME DSLF dataset~\cite{ICME2018}, with disparity ranges between neighboring views $[-2,12]$ and $[-8,6]$, respectively. Both cases contain challenging reflective surfaces, e.g., the ladder in the \textit{DD2} and the metal bottle in the \textit{DD3}. The depth-based view synthesis approach (DIBR~\cite{Acc,CDSD13}) shows tearing artifacts around the occluded thin structures, as shown in the EPIs of the case \textit{DD2}. It also fails to produce reasonable result around the reflective surfaces due to the ambiguous depth in those regions, as shown in the EPIs of the case \textit{DD3}. The results by Wu~\textit{et al.}~\cite{WuEPICNN2018} shows ghosting or blurring effects when addressing the severe aliasing effects in these two cases. Among the baseline methods, the proposed DA$^2$N produces plausible results when reconstructing the LFs with both large disparity and non-Lambertian effect.

Fig. \ref{fig:Result1_2} shows the results ($\times16$ upsampling) of the case \textit{FairyCollection} and \textit{Mannequin} from the MPI Light Field Archive~\cite{kiran2017towards}, with disparity ranges between neighboring views $[-16.5,17]$ and $[-7.5,7]$, respectively. In the demonstrated non-Lambertian materials (EPIs), the disparities are about 14 and 6 pixels, respectively. The significant disparity as well as the non-Lambertian objects (refractive glass) in \textit{Fairycollection} brings extensive challenge for LF reconstruction. Similarly, the specular board in \textit{Mannequin} also suffers non-Lambertian effect. As clearly demonstrated by both the quantitative and qualitative results, the proposed DA$^2$N succeeds to reconstruct the distorted curves by the refraction of the glass. Meanwhile, it can effectively reconstruct not only the white specularity but also the covered red handwriting. The quantitative results on real-world LFs in the MPI Light Field Archive~\cite{kiran2017towards} is listed in Table \ref{tab:result1_2}.


\textit{Ablation study.}  We validate the overall strategies, including the shearing, downscaling and prefiltering operations, and the adopted pseudo EPIs for fine-tuning, by performing the following ablation studies: without the shear layers (``w/o shear'' for short), without the Laplacian pyramid structure for downscaling (``w/o downscale'' for short), without prefilter layers (``w/o prefilter'' for short) and without pseudo EPIs from unstructured LFs for fine-tuning (``w/o pseudo EPIs'' for short). The quantitative results is listed in Table \ref{table:Result1}. Due to the incompleteness of the proposed shearing, downscaling and prefiltering operations for aliasing handling, these approaches show performance degradation in various degrees. Moreover, as indicated by Table \ref{table:Result1}, the performance gap between the network trained on regular LFs (``w/o pseudo EPIs'') and the fine-tuned network (``our proposed'') turns bigger for larger downsampling rate, i.e., from 1.07dB for $\times8$ downsampling rate to 3.96dB for $\times16$ downsampling rate. In other words, the network without fine-tuning on the pseudo EPIs from unstructured LFs suffers a significant performance decrease, especially when addressing severe aliasing.

\begin{figure*}
	\begin{center}
		\includegraphics[width=1\linewidth]{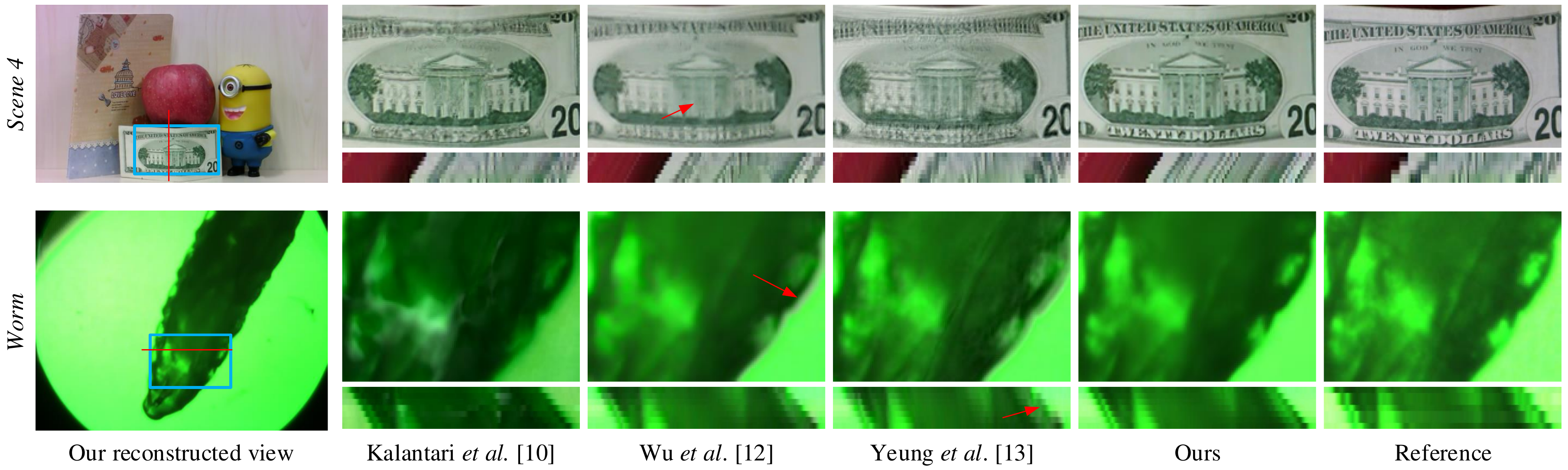}
	\end{center}
	\vspace{-6mm}
	\caption{Comparison of the results on the LFs from camera array system ($\times3$ upsamping). The results show one of the synthesized result, a reference image from the input LF, a portion of the EPI extracted from the location marked in the red line.}
	\label{fig:Result3}
	\vspace{-3mm}
\end{figure*}

\subsection{Light fields from Lytro Illum}
We evaluate the proposed approach using the \textit{30 Scenes} dataset by Kalantari~\textit{et al.}~\cite{DoubleCNN} (contains 30 LFs) , and the \textit{Refractive and Reflective Surfaces} category (contains 32 LFs) as well as the \textit{Occlusions} category (contains 51 LFs) from the Stanford Lytro Light Field Archive~\cite{StanfordLytro}. In this experiment, we reconstruct a $7\times7$ LF from $3\times3$ views ($\times3$ upsampling) and a $8\times8$ LF from $2\times2$ views ($\times7$ upsampling). The baseline methods are two depth independent approaches by Wu~\textit{et al.}~\cite{WuEPICNN2018} and Yeung~\textit{et al.}~\cite{YeungECCV2018}, and two depth-based approaches by Kalantari~\textit{et al.}~\cite{DoubleCNN} and Wu~\textit{et al.}~\cite{wu2019learning}. Since the baseline methods by Wu~\textit{et al.}~\cite{WuEPICNN2018,wu2019learning} used the same structured LFs as the proposed DA$^2$N for training and the other two methods are trained specifically on LFs from Lytro Illum, they are not retrained or fine-tuned in this evaluation.

Fig. \ref{fig:Result2} shows the results of the case \textit{IMG1743} from the \textit{30 Scenes} set~\cite{DoubleCNN} and \textit{Reflective-13} from the \textit{Refractive and Reflective Surfaces} category~\cite{StanfordLytro} using a $3\times3$ input for reconstructing a $7\times7$ LF. The two cases contain large disparities for better demonstrating the aliasing problem. In both cases, the approach by Kalantari~\textit{et al.}~\cite{DoubleCNN} produces tearing artifacts near the occlussion boundaries of the foreground objects, which can be clearly shown in the close-up images and EPIs. The approach by Yeung~\textit{et al.}~\cite{YeungECCV2018} fail to address the aliasing effects around the backgrounds, where the maximum disparity is about 13 pixels. In comparison, our approach provides better results in terms of both local structure (shown in the close-up images) and overall coherency (shown in the EPIs). Table \ref{table:Result2} lists the quantitative results on the evaluated LFs from Lytro Illum. The proposed DA$^2$N achieves the highest average PSNR and SSIM values compared with other approaches. Note that the DA$^2$N is not trained on LFs from Lytro Illum and we do not require additional retraining or fine-tuning for the $\times3$ and $\times7$ upsampling scales.

\textit{Model size and running time.} The model size with respect to number of trainable parameters and running time (with GPU acceleration) for the reconstruction of LFs from Lytro Illum are compared in Table \ref{table:running_time}. It can be shown that the efficiency of our method is comparable as Yeung~\textit{et al.}~\cite{YeungECCV2018}, yet the performance regarding to the aliasing issue of our method is superior as demonstrated in Fig. \ref{fig:Result2}.

\subsection{Light fields from camera array system}
In this section, the camera array-based dataset provided by Wang~\textit{et al.}~\cite{wang2016light} and the microscope dataset provided by Lin~\textit{et al.}~\cite{microLFArray} are evaluated. We reconstruct an LF with $7\times7$ views from a $3\times3$ input ($\times3$ upsampling).

\begin{table}
\caption{Comparison of model size (number of trainable parameters) and running time for the reconstruction of LFs from Lytro Illum.}
\vspace{-5mm}
\begin{center}
\begin{tabular}{lcc}
\toprule
Method & Model size & Running time (s)\\
\hline
Kalantari \textit{et al}.~\cite{DoubleCNN} & 1654004 & 532.5\\
Wu \textit{et al}.~\cite{WuEPICNN2018}  & 56769 & 303.2\\
Yeung \textit{et al}.~\cite{YeungECCV2018}  & 392828 & 6.3\\
Wu \textit{et al}.~\cite{wu2019learning} & 206305 & 88.3\\
Our proposed & 541827 & 19.8\\
\bottomrule
\end{tabular}
\end{center}
\label{table:running_time}
\vspace{-4mm}
\end{table}

Fig. \ref{fig:Result3} shows the reconstructed LFs of the case \textit{Scene 4} from the dataset by Wang~\textit{et al.}~\cite{wang2016light} and the case \textit{worm} from the microscope dataset by Lin~\textit{et al.}~\cite{microLFArray}. The first case~\cite{wang2016light} shows a simple scene with clear occlussion relations and maximum disparity around 15 pixels. The results produced by the baseline methods from Kalantari~\textit{et al.}~\cite{DoubleCNN} and Yeung~\textit{et al.}~\cite{YeungECCV2018} show aliasing effect due to the large disparity, which can be clearly seen in the sub-aperture images and the EPIs in Fig. \ref{fig:Result3}. While the approach by Wu~\textit{et al.}~\cite{WuEPICNN2018} shows blurring artifects due to the high-frequency loss introduced by their ``blur-restoration-deblur'' framework. The second case shows a translucent worm (drosophila larva) captured by a camera array-based microscopy~\cite{microLFArray}, which is challenge due to the large disparity as well as the non-Lambertian. Due to the ambiguous depth information, the approach by Kalantari~\textit{et al.}~\cite{DoubleCNN} fails to render views with reasonable appearance. The result by Wu~\textit{et al.}~\cite{WuEPICNN2018} shows blur effects due to the large disparities. While the approach by Yeung~\textit{et al.}~\cite{YeungECCV2018} produces views with aliasing effects, as shown by the EPI marked with the red arrow in Fig. \ref{fig:Result3}. Among these approaches, our DA$^2$N is able to reconstruct a high quality LF in this non-Lambertian scene.

\begin{figure*}
	\begin{center}
		\includegraphics[width=1\linewidth]{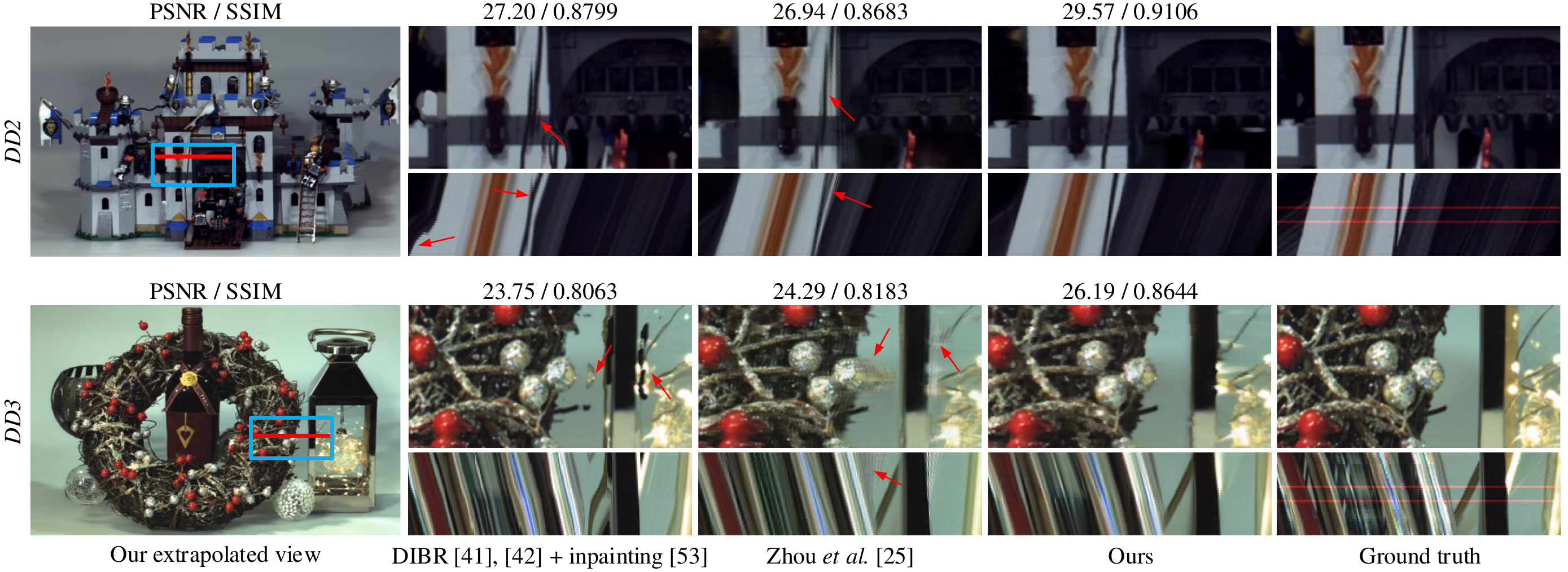}
	\end{center}
	\vspace{-6mm}
	\caption{Comparison of the extrapolation results on the LFs from gantry system. In each case, we applies 2 images (marked in the red lines in the ground truth EPIs) to render a $1\times35$ LF with 7 interpolated views and 28 extrapolated views. The PSNR and SSIM metrics are averaged on the extrapolated views.}
	\label{fig:Result4}
	\vspace{-2mm}
\end{figure*}

\subsection{Application for view extrapolation}
We further extend the proposed LF reconstruction approach for view extrapolation. Different with the LF interpolation, however, the view extrapolation requires the prediction of occluded regions which are hidden from all the input views. In the EPI domain, it is actually closer to the outpainting problem, that is, the network needs to search texture patterns from the input (or reconstructed) EPI and pads them outwards. The proposed DA$^2$N (``shearing-downscaling-prefiltering'' structure) is designed to solve the aliasing problem by reducing disparity (range) in the sampled light field, making it easier for the network to find the correct patterns for view extrapolation. In addition, by combining with a discrimination network~\cite{goodfellow2014generative}, the DA$^2$N is able to inference the extrapolated view and predict the occluded background simultaneously in the EPI domain.

\textit{Implementation.} The overall network architecture for light field view extrapolation is the same as the one we introduced in Sec. \ref{Sec:network}, except that the stride of the ``deconv5'' layer in the reconstruction layer is $[1,2]$. The other difference is that we introduce a discrimination network to construct a generative adversarial network (GAN) together with the proposed DA$^2$N. The discrimination network has a similar structure with that introduced by Iizuka~\textit{et al.}~\cite{iizuka2017globally}, which contains a global discriminator and a local discriminator for both global and local consistency in the EPI domain. Specifically, the output of the generator (DA$^2$N) is fed to the global discriminator, while the synthesized part is fed to the local discriminator. Then a concatenation layer is applied to connect the outputs from the two discriminators in the channel dimension and yields the final discrimination value. Accordingly, we also add a standard adversarial term~\cite{goodfellow2014generative} to the original loss function described in Eqn. \ref{eq:loss} for the generator
\begin{equation}\nonumber
\mathcal{L}_{gen}(\hat{E},E)=\mathcal{L}(\hat{E},E)+\lambda_{adv}\log(f_{dis}(\hat{E})),
\end{equation}
where $\hat{E}$ and $E$ are the extrapolated EPI and its corresponding label, $f_{dis}(\cdot)$ is the discrimination network and $\lambda_{adv}=-5\times10^{-4}$ is a weighing hyperparameter. 

In this application, the network is trained on the regular LFs from the Stanford Light Field Archieve~\cite{StanfordLFdatasets}. Sub-EPIs with size $14\times64$ are extracted as the training labels, where 7 views are used to extrapolate the rest 7 views, i.e., $\times2$ extrapolation. Note that the extrapolation scale is also flexible similar with the view interpolation via cascade. About 1424K examples are extracted from the 11 LFs~\cite{StanfordLFdatasets}. The learning rate is set to $3\times10^{-4}$ for the generator and $6\times10^{-5}$ for the discrimination network. The network converges after 800K steps of backpropagation. Please refer to the supplementary file for more implementation details.

\textit{Results.} We compared the proposed view extrapolation approaches with two baseline methods, a DIBR~\cite{Acc,CDSD13} approach combined with learning-based image inpainting~\cite{iizuka2017globally} for occluded regions and a multiplane image-based view extrapolation approach by Zhou~\textit{et al.}~\cite{zhou2018MPI}. Fig. \ref{fig:Result4} shows the extrapolation results for the LF from the gantry system~\cite{ICME2018}. In each case, an LF with 2 horizontal views (marked by the red lines in the ground truth EPIs in Fig. \ref{fig:Result4}) are applied to synthesize a $1\times35$ LF, where 7 views are from the interpolation and 28 views from the extrapolation, i.e., $\times5$ extrapolation. The extrapolated views produced by the inpainting-based approach~\cite{iizuka2017globally} appear severe distortion around the occluded regions due to the unreliable estimation of the disoccluded background, which can be shown by the EPIs in Fig. \ref{fig:Result4}. While the results produced by the multiplane image-based approach~\cite{zhou2018MPI} show ghosting artifacts due to the inaccurate depth estimation around occlusion boundaries, which is then magnified by the extrapolation. In comparison, the proposed DA$^2$N is able to extrapolate high quality LF in terms of both visual coherency and quantitative measurement (see the PSNR and SSIM values averaged on the extrapolated views in Fig. \ref{fig:Result4}).

\begin{figure}
\begin{center}
\includegraphics[width=1\linewidth]{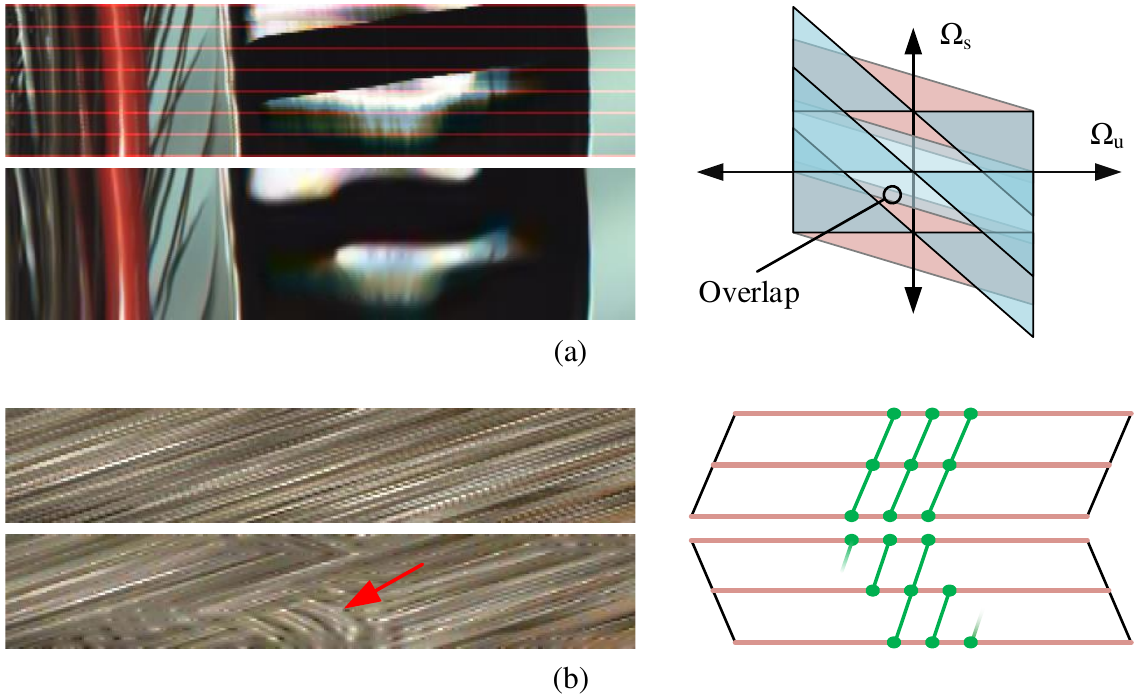}
\end{center}
\vspace{-6mm}
   \caption{Limitations of the proposed method. (a) In a non-Lambertian case with extremely sparse input views (top left), there will be overlap between the expansionary spectra and its replicas (right), which cannot be eliminated through downscaling or prefiltering in the spatial dimension; (b) In some cases with repetitive pattern, the reconstruction net will generate several ``plausible'' outputs using different shear amounts (right), which will lead to ambiguity when being blended together.}
\label{fig:limitation}
\end{figure}

\subsection{Limitations}
A main limitation of the proposed anti-aliasing framework is that we could not reconstruct the non-Lambertian effect that is sampled under an extremely sparse sampling. As demonstrated in Fig. \ref{fig:limitation}(a), when the downsampling rate in the angular dimension is greater than $\frac{\pi Z}{\beta} + 1$, the expansionary spectra will overlap with its replicas (please refer to Eqn. \ref{eq:non-lam} and Fig. \ref{fig:FA}). Unfortunately, this overlap or aliasing part cannot be eliminated through downscaling or prefiltering in the spatial dimension. Another limitation is that the proposed DA$^2$N sometimes fail to reconstruct regions with repetitive pattern. As illustrated in Fig. \ref{fig:limitation}(b), the reconstruction net generates multiple ``plausible'' results using different shear amounts, which lead to ambiguity after being fed into the subsequent fusion net.

\section{Conclusion and Future Work}
Solving the aliasing problem is a key issue for LF reconstruction under the large disparity and non-Lambertian challenges. In this paper, we introduce a deep learning pipeline that is specifically designed for aliasing handling. We analytically show that the internal components of the proposed deep learning pipeline is equivalent to the conventional reconstruction filter in the Fourier domain in solving the aliasing problem. Specifically, the network includes a shear layer to warp the input LF to certain depths, a learning-based Laplacian pyramid to reduce disparity range and a prefilter layer to further remove the aliasing component. To explicitly enhance the capacity on aliasing handling, we begin by training the network from scratch using regular LFs and then fine-tune it with pseudo EPIs from unstructured LFs. We analytically indicate that pseudo EPIs have the same non-Lambertian property in terms of the visual features and the aliasing issue. By performing evaluations on LFs from different acquisition geometries (gantry system, hand-held plenoptic camera and camera array system), we have demonstrated the efficacy and robustness for aliasing handling.

An interesting aspect of future work is to promote the fusion net with flexible input, that is, a variable number of shear layers. This setting will enable users to customize the network architecture according to the disparity range as well as the computational complexity without retraining. To achieve this feature, the fusion net is required to be fully convolutional along the angular, width and shear dimensions, i.e., a 3D network, while keeping only the angular and width dimensions before producing the final EPI.



%

\ifCLASSOPTIONcompsoc
  \section*{Acknowledgments}
\else
\fi
This work was supported by the Major Program of National Natural Science Foundation of China NSFC No.61991400 No.61991401 and No.61991404, Science and Technology Major Projects of Liaoning Province No.2020JH1/10100008, NSFC No.61827805, No.61531014, No.61861166002 and No.6181001011, and Fundamental Research Funds for the Central Universities No. 100802004.

\ifCLASSOPTIONcaptionsoff
  \newpage
\fi



\bibliographystyle{IEEEtran}
\bibliography{IEEEabrv}
%

%

\vspace{-6mm}
\begin{IEEEbiography}[{\includegraphics[width=1in]{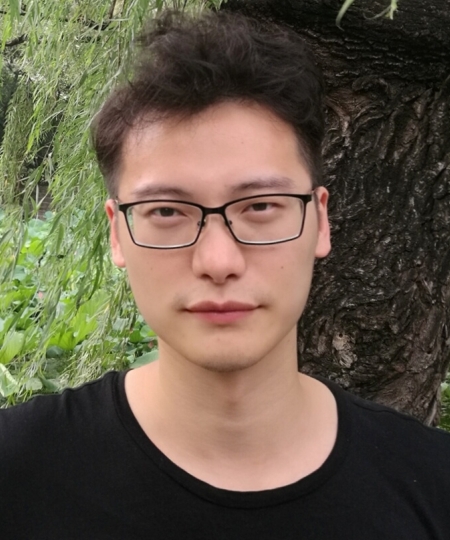}}]{Gaochang Wu}
received the BE and MS degrees in mechanical engineering in Northeastern University, Shenyang, China, in 2013 and 2015, respectively, and Ph.D. degree in control theory and control engineering in Northeastern University, Shenyang, China in 2020. He is currently an associate professor in the State Key Laboratory of Synthetical Automation for Process Industries, Northeastern University. His current research interests include image processing, light field processing and deep learning.
\vspace{-4mm}
\end{IEEEbiography}

\vspace{-2mm}
\begin{IEEEbiography}[{\includegraphics[width=1in]{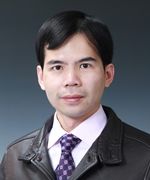}}]{Yebin Liu}
received the BE degree from Beijing University of Posts and Telecommunications, China, in 2002, and the PhD degree from the Automation Department, Tsinghua University, Beijing, China, in 2009. He has been working as a research fellow at the computer graphics group of the Max Planck Institute for Informatik, Germany, in 2010. He is currently an associate professor in Tsinghua University. His research areas include computer vision and computer graphics.
\vspace{-4mm}
\end{IEEEbiography}

\vspace{-2mm}
\begin{IEEEbiography}[{\includegraphics[width=1in]{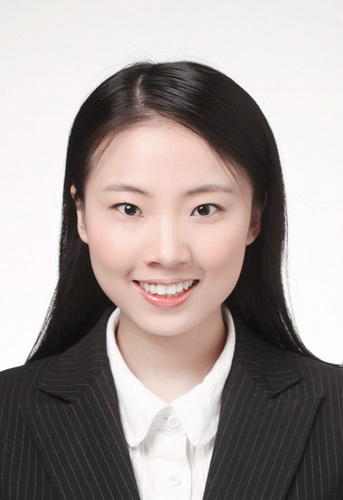}}]{Lu FANG}
is currently an Associate Professor at Tsinghua University. She received her Ph.D in Electronic and Computer Engineering from HKUST in 2011, and B.E. from USTC in 2007, respectively. Dr. Fang's research interests include image / video processing, vision for intelligent robot, and computational photography. Dr. Fang serves as TC member in Multimedia Signal Processing Technical Committee (MMSP-TC) in IEEE Signal Processing Society.
\vspace{-4mm}
\end{IEEEbiography}

\vspace{-2mm}
\begin{IEEEbiography}[{\includegraphics[width=1in]{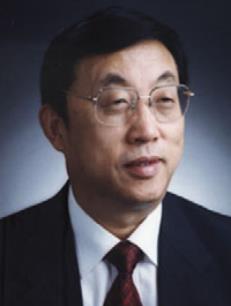}}]{Tianyou Chai}
received the Ph.D. degree in control theory and engineering from Northeastern University, Shenyang, China, in 1985. He has been with the Research Center of Automation, Northeastern University, Shenyang, China, since 1985, where he became a Professor in 1988 and a Chair Professor in 2004. His current research interests include adaptive control, intelligent decoupling control, integrated plant control and systems, and the development of control technologies with applications to various industrial processes. Prof. Chai is a member of the Chinese Academy of Engineering, an academician of International Eurasian Academy of Sciences, IEEE Fellow and IFAC Fellow. He is a distinguished visiting fellow of The Royal Academy of Engineering (UK) and an Invitation Fellow of Japan Society for the Promotion of Science (JSPS).
\end{IEEEbiography}




\end{document}